\documentclass[10pt,twocolumn,letterpaper]{article}

\usepackage{iccv}              

\usepackage[dvipsnames]{xcolor}

\usepackage[accsupp]{axessibility} 

\usepackage{graphicx}
\usepackage{amsmath}
\usepackage{amssymb}
\usepackage{booktabs}
\usepackage{bbm}
\usepackage{multirow}
\usepackage{pifont}
\usepackage{amssymb}
\usepackage{overpic}
\usepackage{makecell}
\usepackage{tabularx}
\usepackage{ragged2e}
\usepackage{afterpage}
\newcolumntype{C}{>{\Centering\hspace{0pt}}X}

\definecolor{cvprblue}{rgb}{0.21,0.49,0.74}
\usepackage[pagebackref,breaklinks,colorlinks,allcolors=cvprblue]{hyperref}

\usepackage[capitalize]{cleveref}
\crefname{section}{Sec.}{Secs.}
\Crefname{section}{Section}{Sections}
\Crefname{table}{Table}{Tables}
\crefname{table}{Tab.}{Tabs.}
 
\def\paperID{6852} 
\def\confName{ICCV}
\def\confYear{2025}

\title{Learning 3D Scene Analogies with Neural Contextual Scene Maps}

\author{Junho Kim\textsuperscript{1}, Gwangtak Bae\textsuperscript{1}, Eun Sun Lee\textsuperscript{1}, and Young Min Kim\textsuperscript{1, 2}
\and {\small \phantom{ }} \vspace{-1em}\\
\textsuperscript{1} {\small Dept. of Electrical and Computer Engineering, Seoul National University} \\
\textsuperscript{2} {\small Interdisciplinary Program in Artificial Intelligence and INMC, Seoul National University} \\
{\tt\small \{82magnolia, tak3452, eunsunlee, youngmin.kim\}@snu.ac.kr}
}

\begin{document}
\twocolumn[{%
\renewcommand\twocolumn[1][]{#1}%
\maketitle
\begin{center}
    \centering
    \captionsetup{type=figure}
    \includegraphics[width=\linewidth]{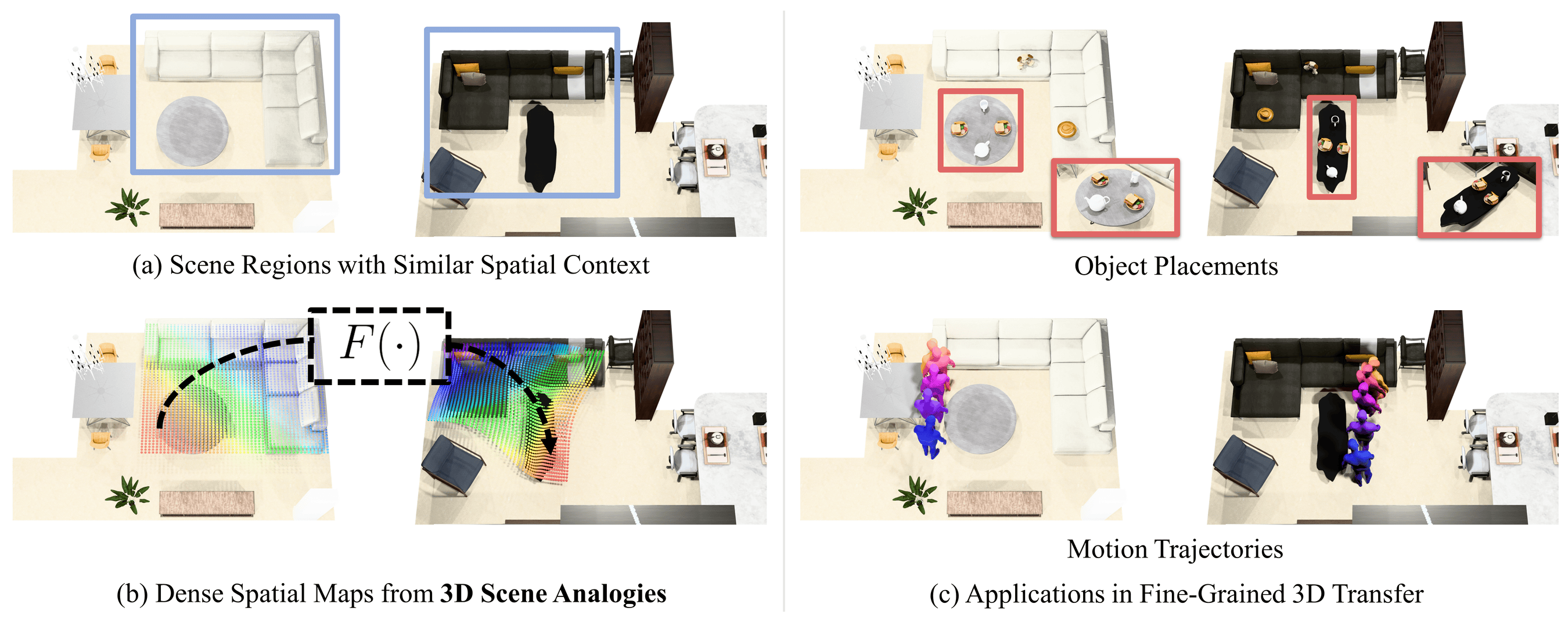}
    \vspace{-2em}
    \captionof{figure}{Overview of the 3D scene analogy task. (a) Given two scenes with regions possibly having similar contexts, (b) the 3D scene analogy task aims to find a dense 3D mapping between the corresponding regions. (c) The estimated maps can then be used for applications such as object placement or motion trajectory transfer.}
   \label{fig:teaser}
\end{center}}]
\begin{abstract}
Understanding scene contexts is crucial for machines to perform tasks and adapt prior knowledge in unseen or noisy 3D environments.
As data-driven learning is intractable to comprehensively encapsulate diverse ranges of layouts and open spaces, we propose teaching machines to identify relational commonalities in 3D spaces.
Instead of focusing on point-wise or object-wise representations, we introduce \textbf{3D scene analogies}, which are smooth maps between 3D scene regions that align spatial relationships.
Unlike well-studied single instance-level maps, these scene-level maps smoothly link large scene regions, potentially enabling unique applications in trajectory transfer in AR/VR, long demonstration transfer for imitation learning, and context-aware object rearrangement.
To find 3D scene analogies, we propose neural contextual scene maps, which extract descriptor fields summarizing semantic and geometric contexts, and holistically align them in a coarse-to-fine manner for map estimation.
This approach reduces reliance on individual feature points, making it robust to input noise or shape variations.
Experiments demonstrate the effectiveness of our approach in identifying scene analogies and transferring trajectories or object placements in diverse indoor scenes, indicating its potential for robotics and AR/VR applications.
Project page including the code is available through this link: \small{\url{https://82magnolia.github.io/3d_scene_analogies/}}.

\vspace{-1.5em}
\end{abstract}
\section{Introduction}
\label{sec:intro}
The 3D world is rich in contextual information, shaped by the interplay of object placements and surrounding open spaces~\cite{scene_context_oliva,scene_context_biederman}.
The function of an object is often flexible, shifting according to its location and spatial relationship to nearby elements; a table might serve as a TV stand in one context or as a tea table beside a sofa in another. 
Capturing these nuanced, high-dimensional relationships is challenging. 
Decades of research in cognitive psychology~\cite{structure_mapping,analogy_1,analogy_2,analogy_3,guarino2022children,scene_analogy_1} suggest that humans rely on analogical reasoning to relate familiar scenes from past experiences to new observations. 
In Figure~\ref{fig:teaser}, humans can intuitively relate areas near a sofa-and-table setup in one room to similar areas in another, yet enabling machines to perform this mapping is far from straightforward. 
To achieve this, one must transfer not only the positions of objects but also their surrounding context which cannot be done through simple object or point-wise matching. 
How can we formulate this problem and extract generalizable representations that encode intricate object relationships and spatial context? 

To address these challenges, we propose the \textbf{3D scene analogy task} of estimating a \textit{dense map} between scenes that share similar contexts, as shown in Figure~\ref{fig:teaser}.
This task demands a smooth map that preserves spatial coherence, allowing consistent relationships across mapped regions without abrupt transitions.
By capturing both individual object placements and their surrounding context, the mapping enables transferring spatial arrangements between scenes in a structure-aware manner. 
This contrasts with conventional feature matching from vision foundation models~\cite{dino,clip,stable_diffusion} or 3D keypoints~\cite{vector_neuron,sprin}, which are often computationally costly or lack scalability for fine-grained scene mapping.
Moreover, these features struggle to capture semantic relationships or nuanced contextual cues necessary for transferring arrangements across scenes~\cite{lerf2023,openscene}.
As such, our task requires a holistic understanding of scene context, allowing for applications where spatial continuity and hierarchical understanding are critical.
One example is in imitation learning for robotics and AR/VR~\cite{teleop,digital_cousins}, where scene-to-scene task transfer can be more practical than generalizing control policies across environments.


Despite its practical benefits, the 3D scene analogy task poses unique challenges not addressed by traditional correspondence methods.
First, a lack of dense ground-truth training data complicates learning, as contextual information varies widely across near-infinite scene configurations.
Second, the task demands holistic reasoning about object relationships and surrounding open spaces at the point level, extending beyond conventional keypoint or scene graph matching methods, which often simplify objects as sparse keypoints or bounding boxes~\cite{semantic_corresp_1,semantic_corresp_2,semantic_corresp_3,semantic_corresp_4,vector_neuron,el2024probing}. 
Finally, robustness to appearance variation is crucial for managing cross-domain differences effectively.

As an effective solution to the 3D scene analogy problem, we introduce neural contextual scene maps.
For a pair of 3D scenes, our method builds descriptor fields that capture detailed spatial relationships and finds matches by aligning the fields using a smooth map.
Input to our method are sparsely sampled scene keypoints and their semantic information, resulting in a lightweight pipeline robust to input variations, noisy geometry, and appearance changes.
Then the descriptor fields gather vicinity information to extract context-aware features.
The fields are trained with contrastive learning, eliminating the need for densely labeled ground-truth data or inductive biases.
Finally, our method estimates a smooth map aligning the descriptor fields through a coarse-to-fine procedure, which reduces the dependence on individual keypoints to reason about the overarching regional relations holistically.


Our approach effectively identifies accurate scene analogies for complex indoor scenes including noisy 3D scans, and is applicable to practical downstream tasks.
Quantitative results show that our method outperforms baselines using vision foundation models~\cite{dino,dinov2,dinov2_1,el2024probing} or scene graphs~\cite{sg_match,sgaligner} on both real and synthetic 3D scenes, despite using a smaller feature dimension and training data.
Additionally, our method also supports mapping \textit{between} real and synthetic scenes indicating its robustness against input domain variations.
We further demonstrate that our pipeline can be used for downstream tasks such as motion trajectory transfer and object placement, which can be extended to transfer long-term demo trajectories for robotics or create co-presence experiences for AR/VR applications. 

To summarize, our main contributions are: 
i) introducing the 3D scene analogy task to find dense mappings between scene regions with common contexts,
ii) developing neural contextual scene maps that combine spatial and semantic contexts of 3D keypoints to create smooth, detailed maps, and
iii) demonstrating our method’s generalizability across various inputs and applications.

\section{Related Work}
\label{sec:related}
\paragraph{Instance and Group Correspondences}
While the 3D scene analogy task is fairly new, there is extensive research on related problems in correspondence estimation, categorized by input settings and granularity.
On the instance level, sparse matching methods in 2D (i.e., \textit{semantic correspondence})~\cite{min2023convolutional, kim2022transformatcher,el2024probing} and 3D~\cite{wang2018learning,park2024learning,vector_neuron} extract neural network features at keypoints to match between instances within the same semantic category.
Similarly, dense matching methods in 2D (also known as \textit{semantic flow})~\cite{kim2017dctm,kim2017fcss, jeon2018parn, kim2018recurrent,gupta2023asic,ju2024robo,semantic_corresp_3,nam2023diffusion} exploit dense features and correlate them in the entire image space for matching.
On the other hand, dense matching methods in 3D~\cite{morreale2024neural,kim2011blended, morreale2021neural,fischer2024nerf} often start by finding sparse correspondences and optimizing smooth surface maps that pass through them.
Our approach extends instance-level dense matching methods to finding \textit{dense maps} over scene regions in 3D sharing similar contexts.

Unlike instance-level, most approaches in group-level correspondence target keypoint or object-wise matches.
In the 2D case, multi-instance semantic correspondence~\cite{lan2021discobox, sun2023misc210k, ren2015multi, xue2011correlative}  aims to link sparse keypoints from an object instance in one image to multiple corresponding instances in another.
For 3D, scene graph matching~\cite{sgaligner,td_ssg,sg_match, miao2025scenegraphloc} seeks correspondences between graphs representing 3D objects as nodes and their relationships as edges~\cite{td_scene_graph_armeni}.
In contrast to these methods focusing on sparse matches, our work finds dense maps of contextually corresponding regions, accounting for both near-surface points and open spaces.
\vspace{-1em}

\paragraph{Neural Fields}
Neural fields are spatio-temporal quantities that are parameterized fully or partially by a neural network~\cite{neural_field_vis_comp}.
Prominent applications of neural fields include photorealistic 3D reconstruction~\cite{mildenhall2021nerf,goli2023nerf2nerf, chen2023mobilenerf}, 3D geometry extraction~\cite{wang2021neus,yariv2020multiview}, and SLAM~\cite{zhu2022nice, wang2023co, wang2022go}.
While these studies primarily focus on visual fidelity and geometric accuracy, more recent works apply neural fields to semantic scene understanding~\cite{fischer2024nerf,hong2022neural,semantic_nerf} and robot motion planning~\cite{simeonov2022neural,wang_attention,sparsedff,dff,clipfields}.
Notably, studies on robot manipulation build fields using features from vision foundation models~\cite{dino,dinov2,dinov2_1,densematcher} for establishing matches between observations during training and deployment.
Our work aims to establish dense, context-aware correspondences that extend beyond visual/geometric fidelity or specific tasks such as manipulation.
Further, while recent works in robotics~\cite{densematcher,dff} consider transfer methods for \emph{single} objects, our work enables transfer between \emph{multiple} objects, encouraging future robotics research on multi-object demonstration transfer.
Utilizing an efficient neural field based on sparse 3D keypoints, we achieve precise matches for both near-surface and open-space regions, which is difficult to attain from existing works.

\section{Method: Neural Contextual Scene Maps}
\label{sec:method}

\begin{figure}[t]
  \centering
    \includegraphics[width=\linewidth]{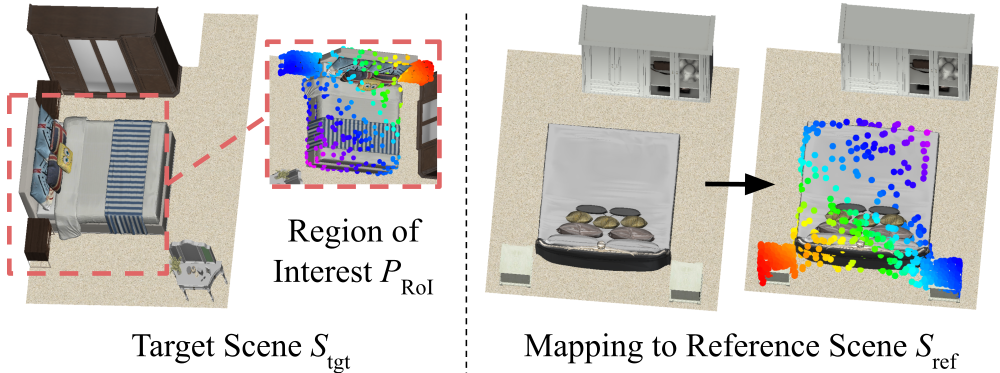}
    \vspace{-1.5em}
   \caption{Overview of our approach. Given a region of interest from object groups in the target scene, our method finds a smooth map to the corresponding region in the reference scene.}
   \vspace{-0.5em}
   \label{fig:overview}
\end{figure}
Given a pair of scenes, our method finds a mapping from a region of interest in one scene to the corresponding region with similar scene contexts in the other scene (Figure~\ref{fig:overview}).
From a sparse set of points sampled in 3D scene models (Section~\ref{sec:input}), our method first builds context descriptor fields that summarize the nearby geometry and semantic information for arbitrary query points (Section~\ref{sec:context_desc}).
Based on descriptor fields, our method finds the dense map in a coarse-to-fine manner, by first extracting an affine map followed by local displacement maps (Section~\ref{sec:map_estimation}).

\subsection{Input Setup}
\label{sec:input}
\paragraph{Region of Interest (RoI) Representation}
Let $S_\text{tgt}$ denote the \textit{target} scene, where the region of interest (RoI) is chosen, and $S_\text{ref}$ the \textit{reference} scene, to which the target scene region is mapped.
As shown in Figure~\ref{fig:overview}, we represent the RoI as a set of points $P_\text{RoI} \subset \mathbb{R}^3$, sampled from the surface of the object group we aim to match.

We then define the \textbf{neural contextual scene map} as a mapping $F(\cdot): \text{conv}(S_\text{tgt}) \rightarrow \text{conv}(S_\text{ref})$, where $\text{conv}(S) \subset \mathbb{R}^3$ denotes the convex hull enclosing scene $S$.
The scene map transforms points $P_\text{RoI}$ to corresponding points in $\text{conv}(S_\text{ref})$ sharing similar scene contexts.
Note, while we use specific points for feature encoding and loss calculation, the final output is a \textit{dense map} across spatial regions, allowing us to find correspondences for any arbitrary point within the region. 
As an illustrative sample,  Figure~\ref{fig:teaser} shows our method mapping between a sofa-and-table group in the target scene to a similar object group in the reference scene.
\vspace{-2em}

\begin{figure}[t]
  \centering
    \includegraphics[width=\linewidth]{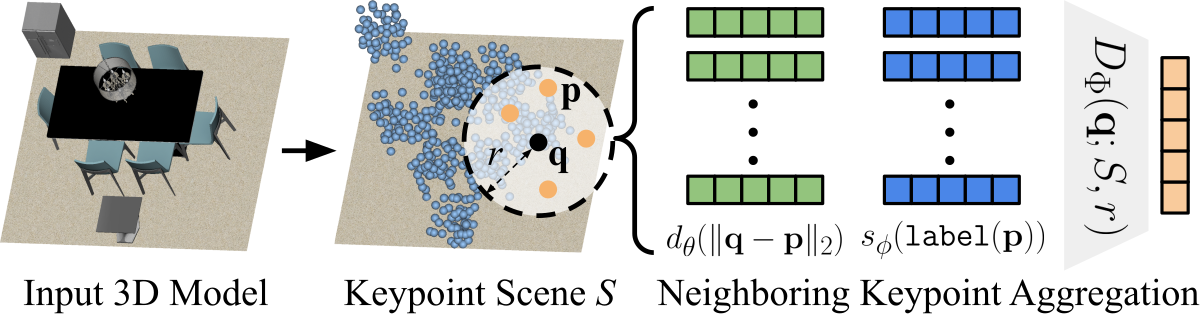}
   \vspace{-1.5em}
   \caption{Overview of context descriptor fields. Using sparsely sampled keypoints as the scene representation, for an arbitrary query point $\mathbf{q}$, the field gathers points within a radius $r$ and computes a Transformer embedding based on the distance embedding $d_\theta(\|\mathbf{q} - \mathbf{p}\|_2)$ and semantic embedding $s_\phi(\mathtt{label}(\mathbf{p}))$.}
   \label{fig:desc_field}
   \vspace{-0.5em}
\end{figure}

\paragraph{Scene Representation} 
As shown in Figure~\ref{fig:desc_field}, our method operates on a lightweight representation of scenes, using sparsely sampled keypoints from the original dense 3D model for efficiency.
Formally, each scene is represented as a tuple $S = (\mathcal{O}, C)$ with an object set $\mathcal{O}$ and scene corner points $C \subset \mathbb{R}^3$.
The object set $\mathcal{O}=\{(P_i, l_i)\}$ consists of points and semantic labels for each object in the scene where $P_i \subset \mathbb{R}^3$ denotes the point coordinates of the $i$\textsuperscript{th} object and $l_i \in \{1, \dots, L\}$ denotes its semantic label among $L$ classes.
Scene corner points are either obtained from floorplan data if available~\cite{td_front,td_future} or from points on convex hulls enclosing the scenes~\cite{convex_hull}.

\subsection{Context Descriptor Fields}
\label{sec:context_desc}
Using sparse input representations, we design descriptor fields as lightweight scene representations that summarize scene context for arbitrary locations by aggregating nearby semantic and geometric information.
For a scene $S$ and a query point $\mathbf{q} \in \text{conv}(S)$, the context descriptor field $D_\Phi(\cdot): \mathbb{R}^3 \rightarrow \mathbb{R}^d$ outputs a $d$-dimensional feature vector.

As shown in Figure~\ref{fig:desc_field}, we implement the descriptor field using a Transformer encoder~\cite{transformer,wang_attention}.
The encoder first aggregates points in $S$ that lie within distance $r$ from $\mathbf{q}$, which we denote as $\mathcal{B}(\mathbf{q}; S, r)$.
For each point $\mathbf{p} \in \mathcal{B}(\mathbf{q}; S, r)$, we concatenate a learned distance embedding $d_\theta(\|\mathbf{q} - \mathbf{p}\|_2)$ and a semantic embedding $s_\phi(\mathtt{label}(\mathbf{p}))$ as Transformer input tokens. 
The descriptor field is defined as follows:
\begin{equation}
    D_\Phi(\mathbf{q}; S, r) = \mathtt{Transformer}(\{\mathtt{Token}(\mathbf{p})\}_{\mathbf{p} \in \mathcal{B}(\mathbf{q}; S, r)}),
    \label{eq:transformer}
\end{equation}
where $\mathtt{Token}(\mathbf{p}) = \mathtt{Concat}(d_\theta(\|\mathbf{q} - \mathbf{p}\|_2), s_\phi(\mathtt{label}(\mathbf{p})))$.
To obtain the feature vector summarizing the input tokens, we append a learnable $\mathtt{[CLS]}$ token to the input token sequence in Equation~\ref{eq:transformer} and use its output embedding as the final field vector~\cite{bert, image_transformer}.

\begin{figure}[t]
  \centering
    \includegraphics[width=\linewidth]{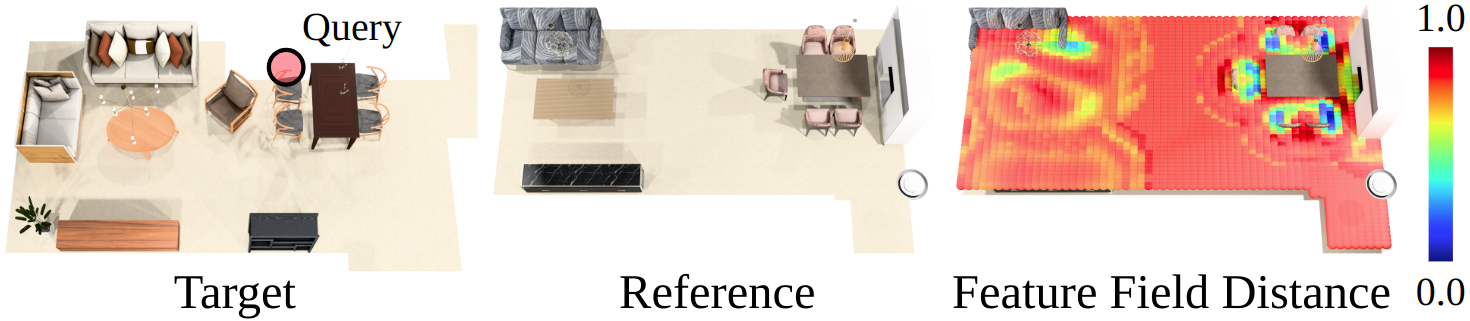}
    \vspace{-2em}
   \caption{Visualization of feature distances in open spaces. For a query point (red) in the target scene lying on the chair arm, we show the descriptor field distances against densely sampled points in the reference scene. Field values are only similar for chair arms near table corners, indicating that descriptor fields can reason about fine-grained contextual correspondences.}
   \label{fig:feature_dist}
   \vspace{-1em}
\end{figure}

Descriptor fields holistically aggregate semantic and geometric information, enabling reasoning about fine-grained contextual correspondences.
As an illustrative sample, Figure~\ref{fig:feature_dist} shows the trained field distances between query points selected at open spaces in the target scene against uniformly sampled points in the reference scene.
Notice sharp peaks are found only near chair arms next to the table corner (and not all chair arms), which indicates that descriptor fields can reason about detailed scene contexts.

\subsubsection{Training Descriptor Fields}
\label{sec:training_process}
To train descriptor fields, we employ contrastive learning~\cite{simsiam,sinclr,sinclrv2,infonce,point_contrast} on procedurally generated positive and negative scene pairs.
Contrastive learning operates by maximizing the similarity of representations for positive data pairs with common attributes while minimizing similarity for negative pairs with dissimilar attributes.
Since contrastive learning only requires positive and negative data pairs~\cite{sinclr,sinclrv2,point_contrast}, our method can learn effective context-aware representations for descriptor fields without densely labeled training data, or hand-tuned inductive biases.
\vspace{-1em}

\begin{figure}[t]
  \centering
    \includegraphics[width=0.95\linewidth]{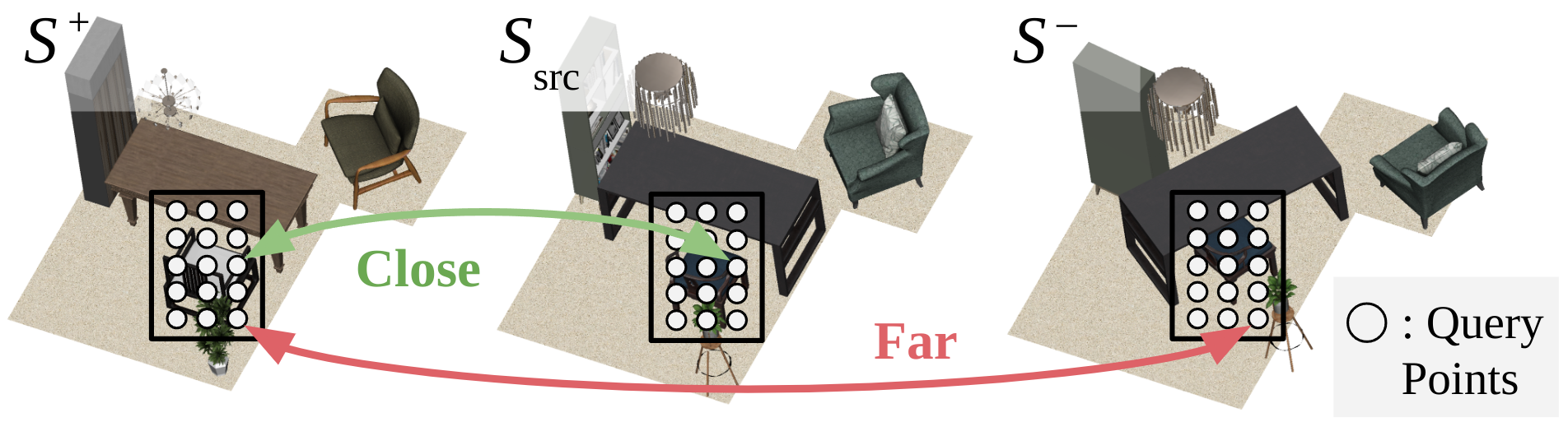}
   \caption{Context descriptor field training overview. We replace each object in scene $S_\text{src}$ with one with the same semantic label to create the positive scene $S^+$, and apply pose noise to obtain $S^-$. Contrastive learning is then applied to descriptor fields computed from points sampled within the object's bounding box.}
   \vspace{-1em}
   \label{fig:desc_train}
\end{figure}

\paragraph{Dataset Generation}
As shown in Figure~\ref{fig:desc_train}, we propose an automated procedure to generate positive and negative scene pairs. 
Our pipeline assumes a \textit{source} dataset consisting of 3D scenes $\mathcal{D}_\text{src} {=} \{S_i\}$ with known object poses.
Among the many possible definitions for a ``correct" correspondence (e.g., appearance~\cite{superglue}, style~\cite{stylerf}, or semantics~\cite{semantic_corresp_1}), we target finding point matches that share common nearby object semantics and local geometry, inspired from works in semantic correspondence~\cite{semantic_corresp_1,semantic_corresp_2}.
Based on this notion, the positive pairs $(S_i, S^+_i)$ are generated by swapping objects in each scene $S_i$ with randomly selected objects sharing the same semantic label from other scenes $\mathcal{D}_\text{src} \setminus S_i$.
Here, the objects for replacement are sampled from the top-K (=100) list of objects having the most similar aspect ratios.
Next, the negative pairs $(S_i, S^-_i)$ are generated by adding noise perturbations to the object poses, similar to LEGO-Net~\cite{legonet}.
Note that we constrain the pose noise to planar translation and z-axis rotation to prevent floating objects or ground penetrations.
The resulting triplet dataset $\mathcal{D}_\text{triplet}{=}\{(S_i,S^+_i,S^-_i)\}$ is used for training.
\vspace{-1em}

\paragraph{Contrastive Learning}
We extract query points from the generated scene triplets $(S_i,S^+_i,S^-_i)$ for contrastive learning.
Specifically, for each object in the source scene $o \in S_i$ and its corresponding object $o^+ \in S^+_i$, we sample an equal number of query points $Q, Q^+$ within the objects' oriented bounding box.
Since objects $o$ and $o^+$ share the same pose, we can associate each positive pair query point $\mathbf{q^+}$ with its corresponding source query point $\mathbf{q} \in Q$, as shown in Figure~\ref{fig:desc_train}.
Setting the negative query points as identical locations to the positive query points, the contrastive learning objective is defined as an InfoNCE loss~\cite{infonce,sinclr,point_contrast} namely,
\begin{equation}
\label{eq:train_objective}
    \mathcal{L}=\sum_{\mathbf{q}, \mathbf{q^+}}-\log \frac{\exp(D_\Phi(\mathbf{q}; S,r)^T D_\Phi(\mathbf{q^+}; S^+,r) / \tau)}{\sum\limits_{\Tilde{S} \in \mathcal{S}}\exp(D_\Phi(\mathbf{q}; S,r)^T D_\Phi(\mathbf{q^+}; \Tilde{S},r) / \tau)},
\end{equation}
where $\mathcal{S}=\{S^+,S^-\}$ and $\tau$ is a temperature parameter set to 0.2 in all our experiments.
Our training objective enforces the descriptor field to output similar embeddings for points lying on positive scene pairs and dissimilar embeddings for those on negative scene pairs.
The trained fields are then used to estimate scene maps in the next section.

\subsection{Contextual Scene Map Estimation}
\label{sec:map_estimation}
\begin{figure}[t]
  \centering
    \includegraphics[width=\linewidth]{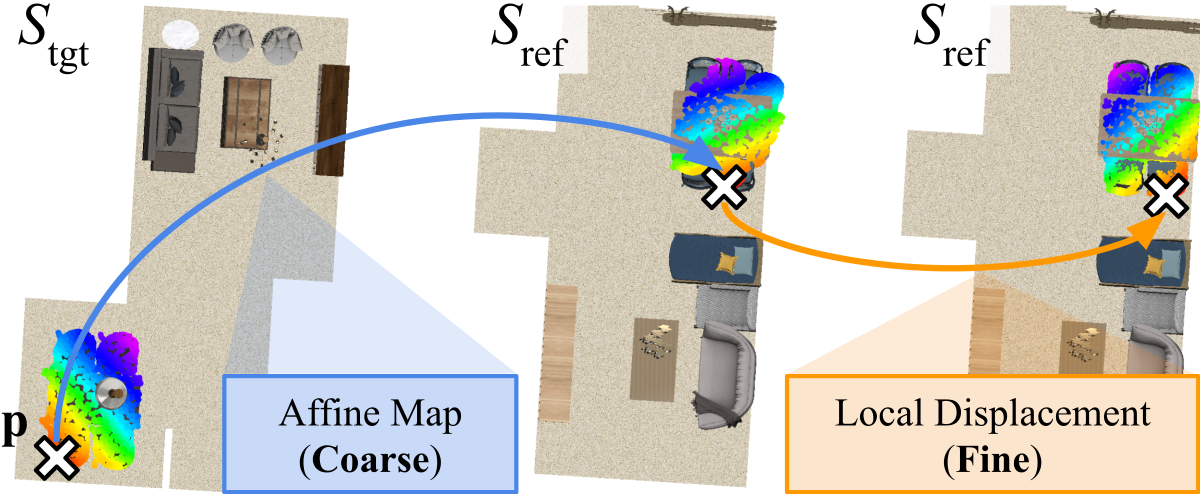}
   \vspace{-1em}
   \caption{Scene map estimation process overview. Our method first estimates affine maps to account for large transformations, and finds local displacements for detailed alignment.}
   \label{fig:scene_map}
   \vspace{-1em}
\end{figure}

We now create a smooth map aligning the descriptor fields between two scenes.
The design intentionally respects spatial vacancies and fine details near keypoints while reducing reliance on individual descriptors for enhanced robustness.
Here, we employ a coarse-to-fine procedure to calculate the contextual map, as shown in Figure~\ref{fig:scene_map}.
Since there are many possible scene arrangements, the target and reference scenes may contain a different number of objects with shape variations.
The coarse initialization with the smooth mapping can effectively ignore minor deviations between the two scenes and focus on deducing a holistic map.
Specifically, we decompose the contextual scene map into an affine map and local displacements,
\begin{equation}
\label{eq:decomp}
    F(\mathbf{x}) := \mathbf{A}\mathbf{x} + \mathbf{b} + d_w(\mathbf{x}; P_\text{RoI}),
\end{equation}
where $\mathbf{A} \in \mathbb{R}^{3\times3}$, $\mathbf{b} \in \mathbb{R}^3$ are the affine map parameters.
We express the local displacement map as a linear combination of radial basis functions~\cite{rbf_interp,rbf_interp_2},
\begin{equation}
\label{eq:displacement}
    d_w(\mathbf{x}; P_\text{RoI}) = \sum_{k} w_k \varphi(\|\mathbf{x} - \mathbf{p}_k\|),
\end{equation}
where the control points are set as points on the RoI $\mathbf{p}_k \in P_\text{RoI}$ described in Section~\ref{sec:input}. 
The basis function is set as the thin plate spline $\varphi(r) := r^2\log(r)$.
Intuitively, the affine map accounts for large, global transformations, and the local displacement map provides a fine-grained alignment for regions with similar contexts.
\vspace{-1em}

\paragraph{Affine Map Estimation}
We extract a pool of affine maps by combinatorially associating object pairs in scenes $S_\text{tgt}$ and $S_\text{ref}$, and optimize the initial maps from descriptor field alignment.
Due to the low-dimensional structure of affine maps and the sparse keypoint representation, we can quickly select maps for further optimization.
For each object pair $(o_\text{tgt}, o_\text{ref})$ with centroids $(\mathbf{c_\text{tgt}}, \mathbf{c_\text{ref}})$, we create a set of affine maps by associating object centroid displacements $\mathbf{c_\text{tgt}} - \mathbf{c_\text{ref}}$ with $N_\text{ortho}$ uniformly sampled rotations and reflections in $SO(2)$.
From the resulting $|\mathcal{O}_\text{tgt}| \times |\mathcal{O}_\text{ref}| \times N_\text{ortho}$ maps, we calculate the following cost function for each affine map $(\mathbf{A}, \mathbf{b})$,
\begin{equation}
\label{eq:coarse_cost}
    \mathcal{C}_\text{coarse} {=} \sum_{\mathbf{p} \in P_\text{RoI}} \!\! \|D_\Phi(\mathbf{p}; S_\text{tgt}) - D_\Phi(\mathbf{A}\mathbf{p} + \mathbf{b}; S_\text{ref})\|,
\end{equation}
where the descriptor fields are compared for points lying on the RoI $P_\text{RoI}$.
Note we have omitted the radius input $r$ for brevity.
As the next step, we select $K_\text{coarse}$ affine maps with the smallest cost values and perform a simple outlier object filtering procedure to remove objects in the RoI that are not matchable to the reference scene, where details are deferred to the supplementary material.
After outlier removal, we optimize each affine map by minimizing the cost in Equation~\ref{eq:coarse_cost} with gradient descent~\cite{sgd,adam}.
\vspace{-1em}

\paragraph{Local Displacement Map Estimation}
We finally refine each affine map $(\mathbf{A_\text{opt}}, \mathbf{b_\textbf{opt}})$ selected from the previous step by further aligning fields with local displacements.
Specifically, we minimize the following cost function
\begin{equation}
\label{eq:fine_cost}
    \mathcal{C}_\text{fine} {=} \sum_{\mathbf{p} \in P_\text{RoI}} \!\! \|D_\Phi(\mathbf{p}; S_\text{tgt}) - D_\Phi(\mathbf{A_\text{opt}}\mathbf{p} + \mathbf{b_\text{opt}} + \delta; S_\text{ref})\|
\end{equation}
where $\delta = d_w(\mathbf{p}; P_\text{RoI})$ is the local displacement defined in Equation~\ref{eq:displacement}.
Similar to affine map estimation, we optimize the basis function weights $w_k$ by minimizing Equation~\ref{eq:fine_cost} with gradient descent.
Finally, our method outputs the mapping with the smallest cost if the cost value is below a designated threshold $\rho_\text{valid}$, or otherwise labels the RoI to be \textit{unmappable} to objects in the reference scene.

\section{Experiments}
\label{sec:experiments}
We evaluate our method for estimating 3D scene analogies on a wide range of 3D scenes (Section~\ref{sec:perf}) and examine applicability in downstream tasks (Section~\ref{sec:app}).

\label{sec:setup}
\vspace{-1em}
\paragraph{Baselines}
As finding 3D scene analogies is a new task, we compare our method against several contrived baselines, which are adaptations of recent 3D scene understanding pipelines~\cite{d3fields,dift,dino,dinov2,sparsedff,sgaligner,el2024probing}.
First, the \textit{scene graph matching} baseline constructs 3D scene graphs~\cite{td_scene_graph_armeni} and matches them via graph matching~\cite{td_ssg} followed by affine map estimation from object centroids.
Next, the \textit{multi-view semantic correspondence} baseline estimates 2D semantic correspondences between image rendering pairs of the input scenes using DINOv2 features~\cite{el2024probing,dinov2,dinov2_1}, and lifts the 2D matches to 3D via back-projection.

The \textit{visual feature field} and \textit{3D point feature field} both generate 3D feature fields similar to our method and apply the map estimation from Section~\ref{sec:map_estimation}.
The visual feature field uses back-projected DINOv2~\cite{dinov2} features from scene renderings~\cite{sparsedff}, while the 3D feature field extracts Vector Neuron~\cite{vector_neuron} features for 3D keypoints and interpolates them for arbitrary queries~\cite{d3fields}.
We discuss further details in the supplementary material.

\vspace{-1em}
\paragraph{Datasets}
We evaluate two diverse indoor scene datasets: synthetic 3D scenes from 3D-FRONT~\cite{td_front} and real 3D scans from ARKitScenes~\cite{arkit} that include object semantic, instance, and pose labels suitable for training and evaluation.
Context descriptor fields are trained separately on each dataset.
We generate 10,000 training triplets for 3D-FRONT~\cite{td_front} using the procedure in Section~\ref{sec:training_process} and 4,498 triplets for ARKitScenes~\cite{arkit} following the standard train/test split.
%
In the absence of densely annotated ground-truth, we prepare two types of evaluation data to assess 3D scene analogies.
\begin{itemize}
    \item \textbf{Procedurally generated scene pairs:} 
    For each scene, we randomly select object groups and procedurally create a new scene containing them.
    Since object poses are known for the generated group matches, we apply the Hungarian algorithm~\cite{hungarian} to obtain pseudo ground-truth maps.
    \item \textbf{Manually collected scene pairs:}
    We collect scene pairs with co-present object groups, along with pairs lacking common object groups to check whether any false positive 3D scene analogies are found.
\end{itemize}
We defer details on evaluation data preparation to the supplementary material.
\vspace{-1em}

\paragraph{Implementation Details}
On both datasets, we extract object keypoints from the dense 3D model using farthest point sampling~\cite{fps}.
Scene corner points are obtained from the floorplan corners for 3D-FRONT~\cite{td_front,td_future}, and from convex hull points for ARKitScenes~\cite{arkit}.
For descriptor fields, we set $r{=}0.75$ and $d{=}256$.
During scene map estimation, we set $N_\text{ortho}{=}16$, $K_\text{coarse}{=}5$, $\rho_\text{valid}{=}1.5$, and optimize scene maps using Adam~\cite{adam} with step size $10^{-3}$.

\begin{table}[t]
    \begin{subtable}{\linewidth}
    \centering
    \resizebox{\linewidth}{!}{
    \begin{tabularx}{1.5\textwidth}{l|CC|CC|CC}
    \toprule
    Metric & \multicolumn{2}{c|}{\makecell{PCP}} & \multicolumn{2}{c|}{\makecell{Bijectivity PCP}} & \multicolumn{2}{c}{\makecell{Chamfer Acc.}} \\ \midrule
    Threshold & 0.25 & 0.50 & 0.25 & 0.50 & 0.15 & 0.20 \\ \midrule
    Scene Graph Matching & 0.26 & 0.42 & 0.29 & 0.47 & 0.32 & 0.48 \\
    Multi-view Semantic Corresp. & 0.10 & 0.20 & 0.14 & 0.21 & 0.62 & 0.86 \\
    Visual Feature Field & 0.50 & 0.66 & 0.52 & 0.61 & 0.81 & 0.86 \\
    3D Point Feature Field & 0.56 & 0.71 & 0.60 & 0.68 & 0.86 & 0.89 \\
    Ours & \textbf{0.76} & \textbf{0.90} & \textbf{0.92} & \textbf{0.94} & \textbf{0.97} & \textbf{0.99} \\
    \bottomrule
    \end{tabularx}
    }
    \caption{Procedurally Generated Scene Pairs}
    \end{subtable}

    \begin{subtable}{\linewidth}
    \centering
    \resizebox{0.8\linewidth}{!}{
    \begin{tabularx}{1.2\textwidth}{l|CC|CC}
    \toprule
    Metric & \multicolumn{2}{c|}{Bijectivity PCP} & \multicolumn{2}{c}{Chamfer Acc.} \\ \midrule
    Threshold & 0.25 & 0.50 & 0.15 & 0.20 \\ \midrule
    Scene Graph Matching & 0.22 & 0.36 & 0.27 & 0.40 \\
    Multi-view Semantic Corresp. & 0.03 & 0.06 & 0.21 & 0.45 \\
    Visual Feature Field & 0.56 & 0.58 & 0.69 & 0.75 \\
    3D Point Feature Field & 0.53 & 0.56 & 0.64 & 0.69 \\
    Ours & \textbf{0.70} & \textbf{0.73} & \textbf{0.71} & \textbf{0.76} \\
    \bottomrule
    \end{tabularx}
    }
    \caption{Manually Collected Scene Pairs}
    \end{subtable}

    \vspace{-0.5em}
    \caption{3D scene analogy comparison in 3D-FRONT~\cite{td_front}.}
    \label{tab:td_front}
    \vspace{-1em}
\end{table}

\subsection{Performance Analysis}
\label{sec:perf}
\paragraph{Metrics}
We use three metrics for quantitative evaluation:
\begin{itemize}
    \item {\textbf{Percentage of Correct Points (PCP)~\cite{semantic_corresp_1,semantic_corresp_2,semantic_corresp_3,semantic_corresp_4}:}}
    This metric is used for procedurally generated scene pairs with pseudo ground-truth annotations.
    For points on the region of interest, the metric is defined as follows, $\text{PCP}(P_\text{RoI}) {=}1/|P_\text{RoI}| \sum_{\mathbf{p_\text{RoI}} \in P} \mathbbm{1}[\|F(\mathbf{p}) - \mathbf{p_\text{gt}}\| {\leq} \alpha]$, where $\alpha$ is a threshold parameter.
    \item {\textbf{Bijectivity PCP~\cite{morreale2024neural,morreale2021neural}:}}
    After computing an inverse scene map $F^{-1}(\cdot): S_\text{ref} \rightarrow S_\text{tgt}$ taking $F(P_\text{RoI})$ as input, this metric is defined as follows: $\text{Bi-PCP}(P_\text{RoI}) {=}1/|P_\text{RoI}| \sum_{\mathbf{p} \in P_\text{RoI}} \mathbbm{1}[\|F^{-1} {\circ} F(\mathbf{p}) {-} \mathbf{p}\| {\leq} \alpha]$.
    \item{\textbf{Chamfer Accuracy:}}
    The metric is defined as the percentage of predictions where i) the Chamfer distance~\cite{chamfer_1,chamfer_2} between mapped points $F(P_\text{RoI})$ and sampled points in $S_\text{ref}$ is below a threshold, or ii) no mappings are output for scene pairs with no common object groups.
\end{itemize}
The PCP and Bi-PCP metrics measure point-level accuracy of the estimated maps, while Chamfer accuracy evaluates group-level accuracy and penalizes false positive maps.

\begin{figure}[t]
  \centering
    \includegraphics[width=0.95\linewidth]{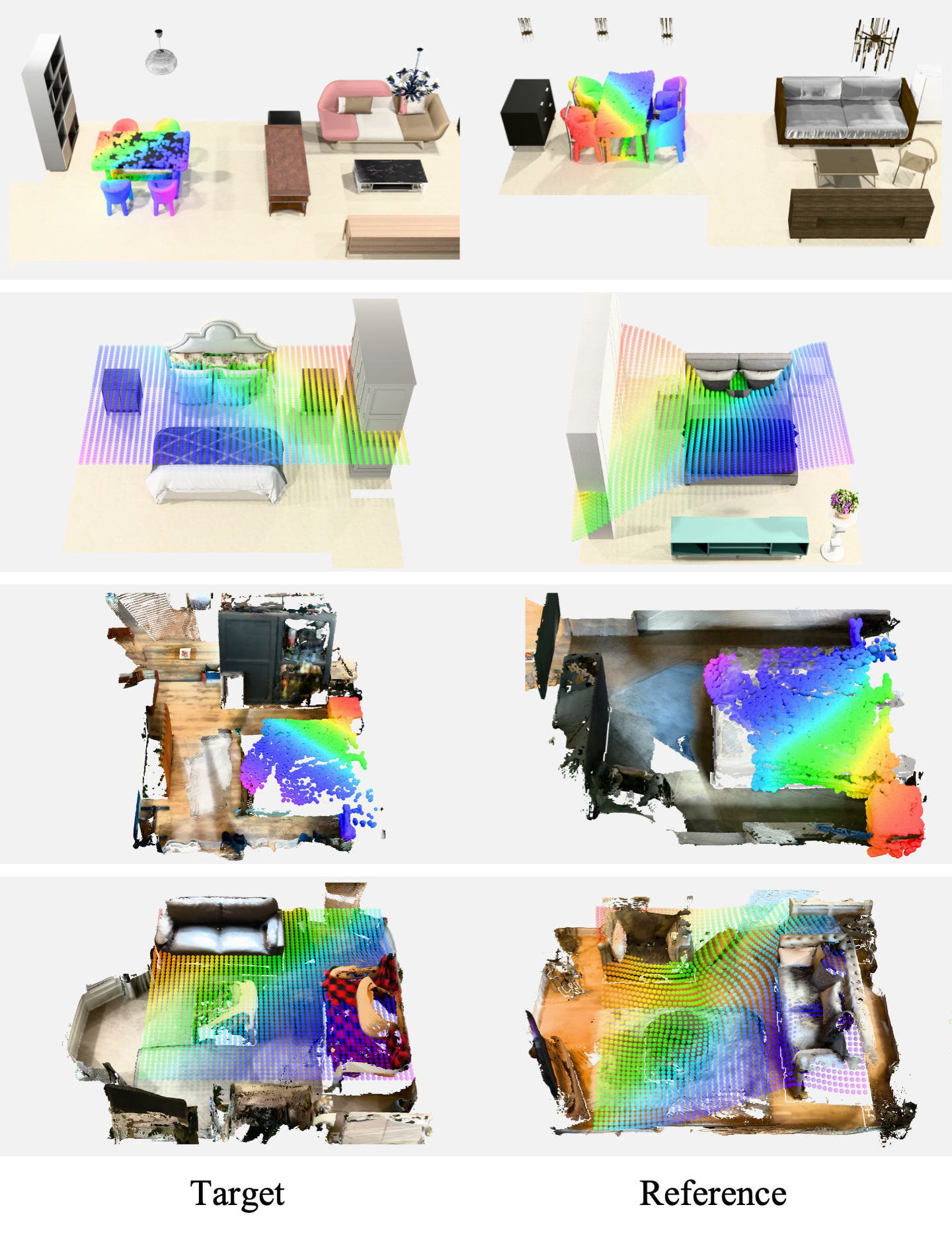}
    \vspace{-1em}
   \caption{Visualizations of estimated 3D scene analogies in 3D-FRONT and ARKitScenes. We show mapping results both for near-surface and open-space points.}
   \vspace{-1em}
   \label{fig:qualitative_scene_map}
\end{figure}
\subsubsection{Scene Map Evaluation}
\paragraph{3D-FRONT}
We first present a quantitative comparison of the scene analogies found from our method with those from baselines on 3D-FRONT~\cite{td_front} in Table~\ref{tab:td_front}.
Our method consistently outperforms the baselines across all metrics for both procedurally generated and manually collected pairs.
A large performance gap exists compared to the scene graph matching baseline, as it treats objects as single nodes, lacking geometric granularity.
A similar trend is observed with the multi-view semantic correspondence baseline.
While recent semantic correspondence methods excel at single object matching~\cite{semantic_corresp_1,semantic_corresp_2,el2024probing}, they struggle to account for spatial relationships among multiple objects. 
Further, the feature field baselines based on DINOv2~\cite{dinov2} and Vector Neurons~\cite{vector_neuron} also exhibit lower performance compared to our method, despite using the same coarse-to-fine map estimation process.
Our method's contrastive learning pipeline enables effective descriptor extraction for highly accurate mappings, as shown in Figure~\ref{fig:qualitative_scene_map}.
\vspace{-1em}

\paragraph{ARKitScenes}
We conduct further assessments on ARKitScenes~\cite{arkit}, which, unlike 3D-FRONT, contains 3D scene meshes from real-world RGB-D camera measurements with noisy geometry and object layouts.
As shown in Table~\ref{tab:arkit}, our method outperforms baselines on most metrics, similar to 3D-FRONT~\cite{td_front}, and generates accurate mappings as shown in Figure~\ref{fig:qualitative_scene_map}.
This indicates that our descriptor fields and coarse-to-fine mapping scheme robustly handle the noisy inputs from real 3D scans.
Nevertheless, all metrics show a consistent performance drop compared to the 3D-FRONT~\cite{td_front} results in Table~\ref{tab:td_front} for manually collected scene pairs.
We attribute this drop to largely incomplete geometry in several manually split scenes, which could be solved by modifying the cost functions in Equation~\ref{eq:coarse_cost},~\ref{eq:fine_cost} to account for such outliers.
While such a level of incompleteness is uncommon in real-world applications,  addressing this issue  is left for future work. 
\vspace{-1em}

\begin{figure}[t]
  \centering
    \includegraphics[width=0.95\linewidth]{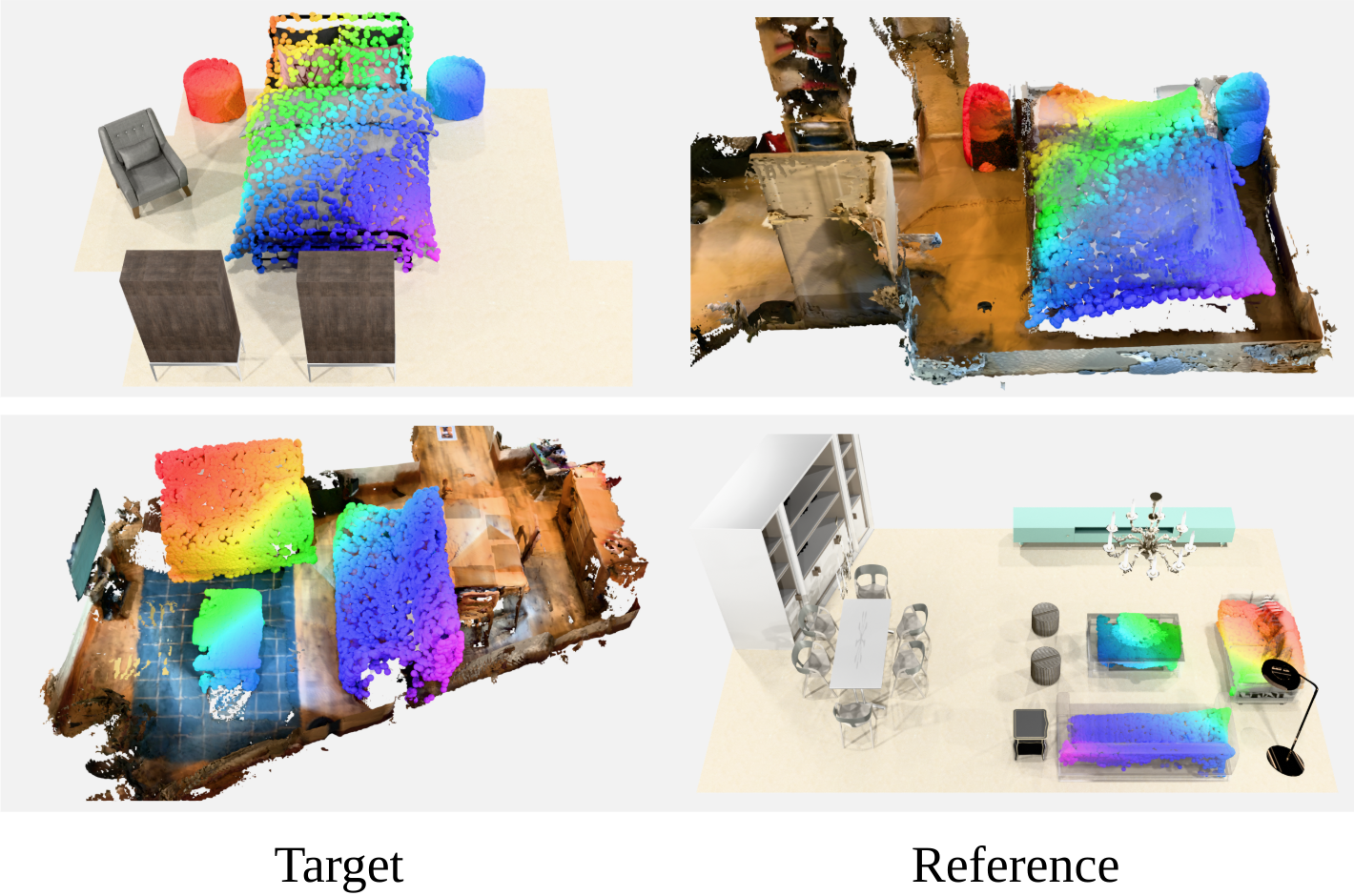}
    \vspace{-1em}
   \caption{Visualizations of Sim2Real and Real2Sim scene analogies estimated between 3D-FRONT and ARKitScenes.}
   \vspace{-1em}
   \label{fig:sim2real}
\end{figure}

\paragraph{Sim2Real and Real2Sim Map Estimation}
We investigate if our method can find analogies between synthetic 3D models in 3D-FRONT~\cite{td_front} and real scans in ARKitScenes~\cite{arkit}.
Such capability is valuable for robotics and AR/VR applications: transferring pre-trained robot policies from virtual simulators to the real world~\cite{digital_cousins}, or enabling immersive telepresence by mapping real-world objects to their virtual counterparts~\cite{gradual_reality}.
Figure~\ref{fig:sim2real} shows estimated scene analogies for both sim-to-real and real-to-sim scenarios, using descriptor fields trained on 3D-FRONT~\cite{td_front} in both cases. 
The coarse-to-fine process allows holistic scene mapping, avoiding over-focus on individual descriptors and achieving reliable mappings across different domains. 
Additional results are in the supplementary material. 

\begin{table}[t]
    \begin{subtable}{\linewidth}
    \centering
    \resizebox{\linewidth}{!}{
    \begin{tabularx}{1.5\textwidth}{l|CC|CC|CC}
    \toprule
    Metric & \multicolumn{2}{c|}{PCP} & \multicolumn{2}{c|}{Bijectivity PCP} & \multicolumn{2}{c}{Chamfer Acc.} \\ \midrule
    Threshold & 0.25 & 0.50 & 0.25 & 0.50 & 0.15 & 0.20 \\ \midrule
    Scene Graph Matching & 0.39 & 0.57 & 0.43 & 0.62 & 0.57 & 0.72 \\
    Multi-view Semantic Corresp. & 0.10 & 0.21 & 0.10 & 0.18 & 0.59 & 0.78 \\
    Visual Feature Field & 0.55 & 0.74 & 0.58 & 0.71 & 0.91 & 0.88 \\
    3D Point Feature Field & 0.65 & 0.81 & 0.70 & 0.77 & 0.88 & 0.92 \\
    Ours & \textbf{0.75} & \textbf{0.90} & \textbf{0.90} & \textbf{0.94} & \textbf{0.96} & \textbf{0.99} \\
    \bottomrule
    \end{tabularx}
    }
    \caption{Procedurally Generated Scene Pairs}
    \end{subtable}

    \begin{subtable}{\linewidth}
    \centering
    \resizebox{0.8\linewidth}{!}{
    \begin{tabularx}{1.2\textwidth}{l|CC|CC}
    \toprule
    Metric & \multicolumn{2}{c|}{Bijectivity PCP} & \multicolumn{2}{c}{Chamfer Acc.} \\ \midrule
    Threshold & 0.25 & 0.50 & 0.15 & 0.20 \\ \midrule
    Scene Graph Matching & 0.25 & 0.37 & 0.33 & 0.45 \\
    Multi-view Semantic Corresp. & 0.06 & 0.12 & 0.31 & 0.50 \\
    Visual Feature Field & 0.26 & 0.29 & 0.40 & 0.42 \\
    3D Point Feature Field & 0.41 & 0.49 & 0.51 & 0.60 \\
    Ours & \textbf{0.51} & \textbf{0.62} & \textbf{0.59} & \textbf{0.69} \\
    \bottomrule
    \end{tabularx}
    }
    \caption{Manually Collected Scene Pairs}
    \end{subtable}

    \vspace{-0.5em}
    \caption{3D scene analogy comparison in ARKitScenes~\cite{arkit}.}
    \label{tab:arkit}
    \vspace{-1em}
\end{table}

\subsubsection{Ablation Study}
\label{sec:abl}
\paragraph{Compatibility with Vision and Language Foundation Models}
We assess our method's compatibility with vision and language foundation model features~\cite{clip,dinov2,bert,sentence_bert} by training variants of the context descriptor fields using CLIP~\cite{clip} or sentence embedding~\cite{sentence_bert} in place of the semantic embedding described in Section~\ref{sec:context_desc}.
CLIP features are extracted from frontal view renders of each object in 3D-FRONT, while sentence embeddings are obtained by captioning each object renders with a vision-language model~\cite{moondream} and extracting text embeddings~\cite{sentence_bert}.
Table~\ref{tab:ablation} shows scene map accuracy for manual and procedural scene pairs, with performance comparable to the original semantic embeddings and outperforming all the baselines.
This shows that our method can effectively incorporate foundation model features without explicit semantic labels.
\vspace{-1em}

\paragraph{Local Displacement Maps}
We finally ablate the coarse-to-fine mapping procedure by comparing our method to a variant that omits the local displacement estimation process.
This results in suboptimal performance, as reported in Table~\ref{tab:ablation}.
Since mappings between scene regions with common contexts are often non-linear, relying solely on the affine map incurs inaccurate scene analogy detections.

\subsection{Applications}
\label{sec:app}
\begin{table}[t]
    \centering
    \resizebox{0.85\linewidth}{!}{
    \begin{tabularx}{.64\textwidth}{l|CC|CC}
    \toprule
    Metric & \multicolumn{2}{c|}{Bijectivity PCP} & \multicolumn{2}{c}{Chamfer Acc.} \\ \midrule
    Threshold & 0.25 & 0.50 & 0.15 & 0.20 \\ \midrule
    Ours w/ CLIP Emb.~\cite{clip} & 0.77 & 0.81 & 0.91 & \textbf{0.97} \\
    Ours w/ Sentence Emb.~\cite{sentence_bert} & 0.78 & 0.82 & 0.92 & \textbf{0.97} \\
    Ours w/o Local Displacement & 0.83 & 0.89 & 0.77 & 0.85 \\
    Ours & \textbf{0.90} & \textbf{0.92} & \textbf{0.94} & 0.96 \\
    \bottomrule
    \end{tabularx}
    }
    \vspace{-0.5em}
    \caption{Ablation study of neural contextual scene maps, averaged on manual and procedural scene pairs from 3D-FRONT~\cite{td_front}.}
    \vspace{-1em}
    \label{tab:ablation}
\end{table}

\begin{figure}[t]
  \centering
    \includegraphics[width=0.95\linewidth]{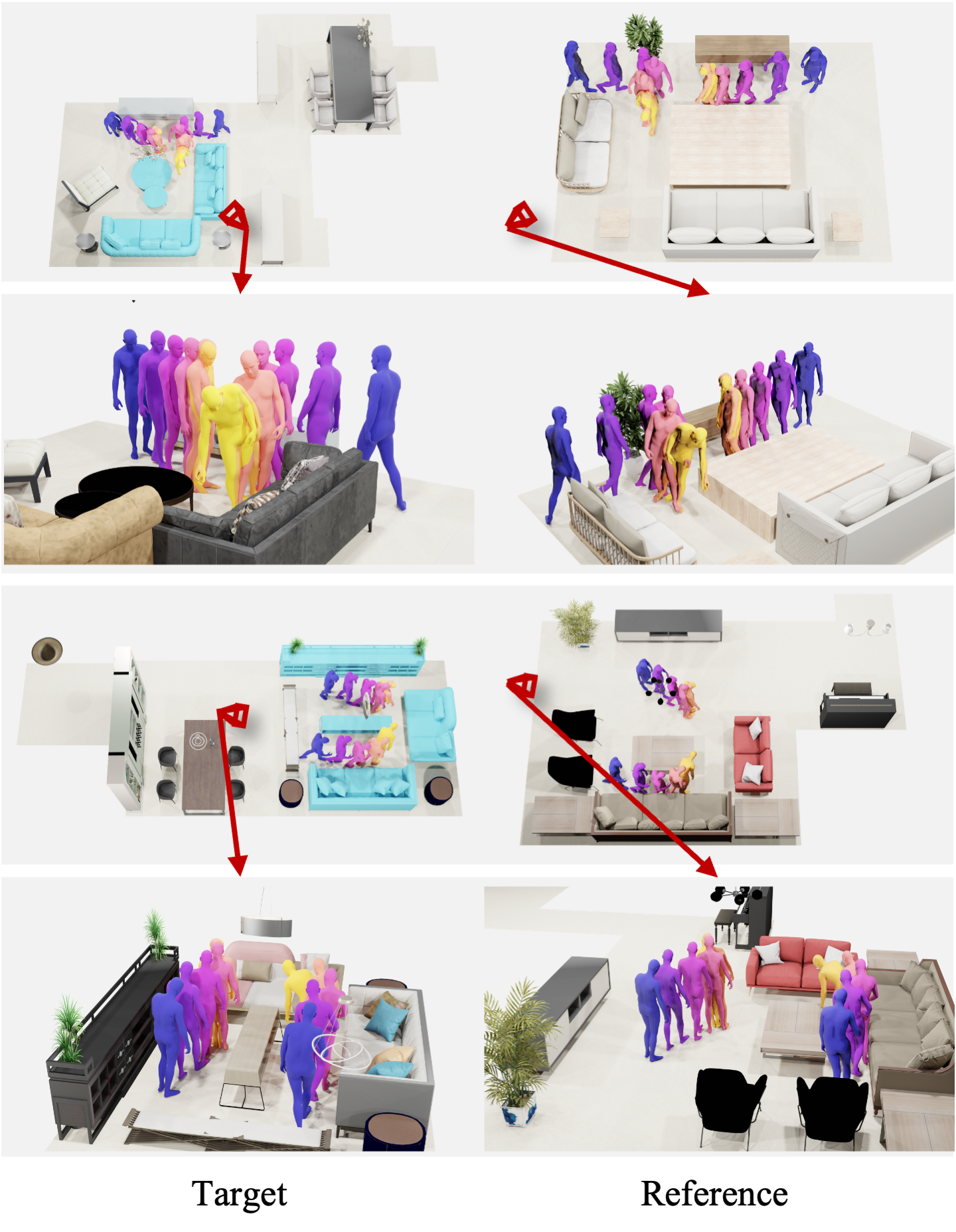}
    \vspace{-1em}
   \caption{Visualization of short trajectory transfer by directly mapping trajectory points. We shade the region of interest in blue.}
   \label{fig:trajectory_transfer}
   \vspace{-1em}
\end{figure}

\begin{figure}[t]
  \centering
    \includegraphics[width=\linewidth]{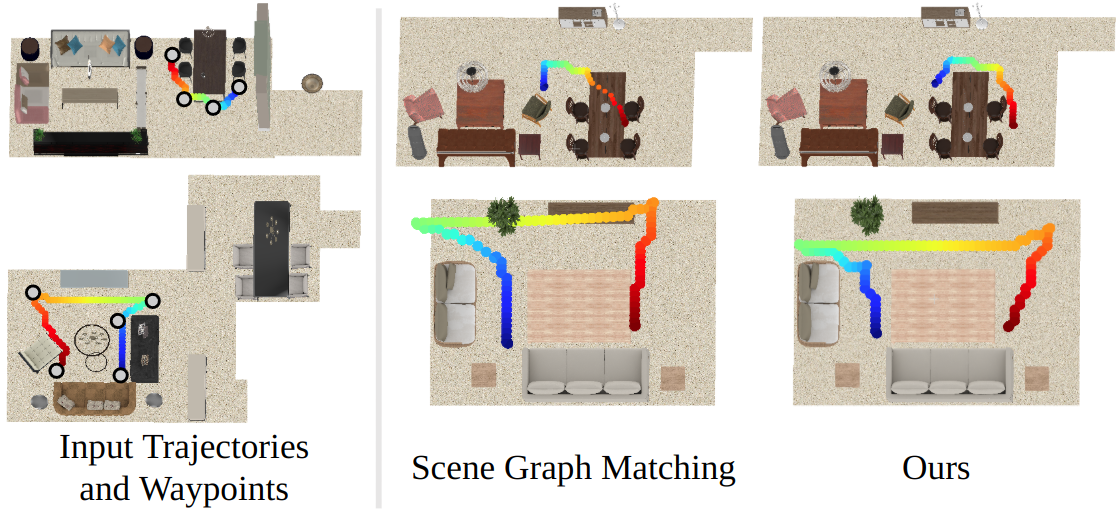}
    \vspace{-1.3em}
   \caption{Comparison of long trajectory transfer against scene graph matching. We estimate scene analogies to map waypoints, and apply traditional path planning~\cite{astar} for interpolation.}
   \label{fig:waypoint_transfer}
   \vspace{-1em}
\end{figure}

\paragraph{Trajectory Transfer}
Given a trajectory in an open space near the region of interest, we test if our method can transfer it to the reference scene's corresponding space.
Such trajectory transfers aid in  teleoperation~\cite{teleop}, data augmentation / demonstration transfer for robot imitation learning~\cite{mimicgen,densematcher}, or virtual co-presence~\cite{copresence} by mirroring the user's trajectory in a virtual environment.
Our method can be applied flexibly depending on the length of the input trajectory.
For short trajectories, we \textit{directly} use the estimated map to transfer each trajectory point.
Figure~\ref{fig:trajectory_transfer} visualizes a short trajectory transfer of virtual human agents moving through the target scene.
We map the virtual human's bounding box corners at each timestamp and use the Umeyama algorithm~\cite{umeyama} to find a rigid transformation to the reference scene.
This result maintains consistent spatial relations with the surrounding objects: for example in Figure~\ref{fig:trajectory_transfer} the virtual human walking between a sofa and a table is accurately transferred.
For long trajectories, directly using the estimated maps may cause collisions.
We integrate our method with classical path planning ~\cite{astar, rrt} by transferring sparse \textit{waypoints} and finding collision-free paths using the A* algorithm~\cite{astar} on the transferred waypoints.
Figure~\ref{fig:waypoint_transfer} shows a comparison against scene graph matching.
We interpolate object surface matches from the baseline to open space~\cite{tps_1,tps_2} to find waypoint transfers, and directly apply the map when the A* algorithm fails due to inaccurate waypoint transfer.
Compared to our method, this process results in erroneous transfers with penetrations.
By producing a smooth map over $\mathbb{R}^3$, our method can flexibly handle trajectory transfer in open spaces, which is difficult with existing pipelines~\cite{sgaligner,object_match_iro,el2024probing} that lack fine-grained understanding of spatial relations and surrounding context.
Additional comparisons are in the supplement.
\vspace{-1em}

\begin{figure}[t]
  \centering
    \includegraphics[width=0.95\linewidth]{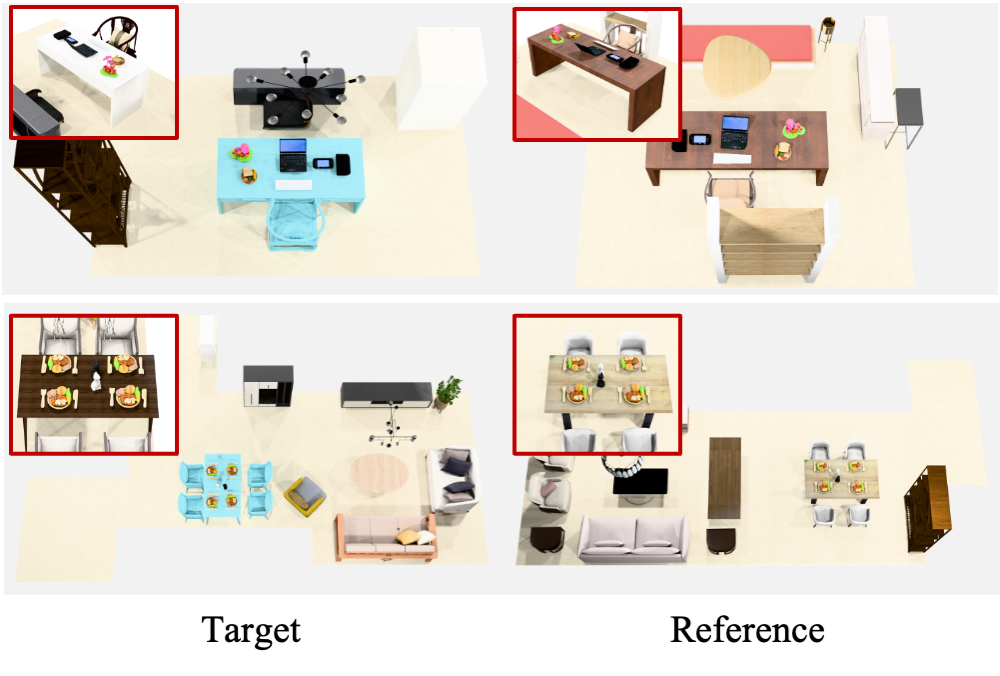}
    \vspace{-1em}
   \caption{Visualization of object placement transfer. We shade the region of interest used for scene analogy estimation in blue.}
   \label{fig:object_placement_transfer}
   \vspace{-1.2em}
\end{figure}

\paragraph{Object Placement Transfer}
In contrast to trajectory and waypoint transfer, which focuses on \textit{open space}, object placement transfer involves mapping small objects placed on a region of interest \textit{surfaces} to the target scene.
We first estimate scene maps from the region of interest and transfer objects placed on its surface via the scene map. 
The task is useful in AR/VR scenarios where users in different physical locations collaborate in a shared virtual space, allowing tools and objects in each user's space to align within the common virtual environment~\cite{gradual_reality,copresence,copresence1,copresence2,copresence3,copresence4}.
As shown in Figure~\ref{fig:object_placement_transfer}, a desk with small items can be accurately mapped from the target to the reference space.
Our method successfully transfers objects to coherent matching locations, demonstrating flexibility in handling both near-surface and open-space transfers.


\vspace{-0.5em}
\section{Conclusion}
\vspace{-0.5em}
We introduce \textit{3D scene analogies}, which are dense maps between scenes with similar contexts, and propose neural contextual scene maps to find smooth, coherent mappings.
Our method uses contextual descriptor fields and an effective coarse-to-fine estimation that holistically aligns the fields.
Experiments demonstrate robustness across real-world scans and sim-to-real scenarios, with applications in trajectory and object placement transfer. 
We hope our work inspires future research in 3D scene context understanding.
\vspace{-1.5em}

\paragraph{Limitations}
We acknowledge several limitations that invite future work.
Currently, our method outputs a single mapping, whereas generating multiple plausible mappings could better account for symmetries and multi-modal correspondences.
Additionally, while our method handles a wide range of spatial variations, it may struggle when object positions swap, as such changes disrupt the initial affine mapping.
Future work could explore more flexible alignment strategies to address these cases.
Finally, our evaluation is based on semantic and local geometric similarity, but different tasks may require alternative notions of ``correctness."
Expanding the evaluation framework to incorporate task-specific criteria could provide deeper insights.
Further details are in the supplementary material.

\paragraph{Acknowledgements}
This work was supported by the National Research Foundation of Korea(NRF) grant (No. RS-2023-00218601) and Institute of Information \& communications Technology Planning \& Evaluation (IITP) grant (No.RS-2021-II212068, Artificial Intelligence Innovation Hub) funded by the Korea government(MSIT), and the BK21 FOUR program of the Education and Research Program for Future ICT Pioneers, Seoul National University in 2025.

\appendix
\renewcommand\thetable{\thesection.\arabic{table}}    
\setcounter{table}{0}
\renewcommand\thefigure{\thesection.\arabic{figure}}    
\setcounter{figure}{0}
\setlength{\tabcolsep}{3pt}
\renewcommand{\theequation}{\thesection.\arabic{equation}}

\setcounter{page}{1}
\section{Method Details}
\subsection{Contextual Descriptor Fields}
\label{sec:supp_context_desc}
\paragraph{Network Architecture}
Contextual descriptor fields described in Section~\textcolor{cvprblue}{3.2} gather semantic and geometric information near query locations to summarize scene context information.
The contextual descriptor field consists of 6 transformer encoder layers~\cite{transformer,bert}, and each encoder layer contains multi-head attention models with 8 heads.
For the semantic embedding, we use a learnable embedding of size 32, and for the distance embedding, we use a multi-layer perceptron (MLP) with a single hidden layer to produce an embedding of size 32.
In addition, the descriptor fields operate on a lightweight scene presentation, where we sample 50 points per object in the scene point cloud using farthest point sampling~\cite{fps}.

\paragraph{Training}
To train descriptor fields, we build a dataset consisting of scene triplets for contrastive learning.
For obtaining positive pairs, we replace each object in the source scene with a randomly sampled object from another scene sharing the same semantic labels.
For negative pairs, we add translation noise sampled from the uniform distribution $\mathcal{U}(-0.5, 0.5)$ and z-axis rotation noise with rotation angles sampled from $\mathcal{U}(-90^\circ, 90^\circ)$.
In addition, we uniformly sample $20^3$ grid points from each object bounding box and an equal number of points sampled near the object surface as query points for training.
For each object pair, we associate each grid point via nearest neighbor matching and surface point via Hungarian matching~\cite{hungarian}.
Then, during training, we minimize the contrastive learning objective (Equation~\textcolor{cvprblue}{2}) using the Adam~\cite{adam} optimizer with a learning rate $10^{-4}$ and query point batch size 4 for $10^4$ epochs.

\subsection{Affine Map Estimation}
\label{sec:affine_supp}
As explained in Section~\textcolor{cvprblue}{3.3}, we estimate large, global transformations using affine maps.
First, we combinatorially associate object pairs in scenes $S_\text{tgt}$ and $S_\text{ref}$ to extract initial sets of affine maps.
Namely, the set of affine maps is created by associating $N_\text{ortho}$ uniformly sampled transforms in $SO(2)$ with translations between object pairs $(o_\text{tgt}, o_\text{ref})$ with centroids $(\mathbf{c_\text{tgt}}, \mathbf{c_\text{ref}})$.
This can be formally expressed as follows, $\{(A_\text{init}, b_\text{init})\} := \{(T_\text{ortho}, -T_\text{ortho}\mathbf{c_\text{tgt}} + \mathbf{c_\text{ref}})\}$, where $T_\text{ortho} \in SO(2)$.
Here, we set $N_\text{ortho}$ by combinatorially associating $N_\text{rot}=4$ uniformly sampled rotations with $N_\text{rfl}=2$ reflections along the x and y axes.
From the initial set, we select $K_\text{coarse}$ affine maps with the smallest cost specified in Equation~\textcolor{cvprblue}{5}.

\paragraph{Outlier Object Rejection}
Prior to gradient descent optimization, we perform a simple filtering procedure to remove outlier objects in the region of interest, i.e., objects that cannot be matched to the reference scene.
For each selected affine map, we identify object instance matches between the region of interest and the reference scene.
To elaborate, we create a distance matrix $D \in \mathbb{R}^{N_\text{RoI} \times N_\text{Ref}}$, where $N_\text{RoI}$ and $N_\text{Ref}$ are number of object instances in the region of interest and reference scene respectively.
The $(i, j)^\text{th}$ entry of the distance matrix is initially set as the point cloud centroid distance between the $i^\text{th}$ object in the region of interest after affine map warping and the $j^\text{th}$ object in the reference scene.
We then assign infinity values to matrix entries where the object semantic labels disagree.
Finally, we perform Hungarian matching on $D$ to find object instance matches, and identify objects in the region of interest as outliers if the distance matrix value to the matched object is over a threshold (which is set to 2.0 in all our experiments).
After outlier removal, we keep affine maps with the largest number of inlier objects and optimize each affine map with gradient descent.

\subsection{Local Displacement Map Estimation}
Given the estimated set of affine maps, our method finds local displacement maps for fine-grained scene context alignments.
To obtain local displacement maps, we first minimize Equation~\textcolor{cvprblue}{6} with gradient descent, treating each local displacement $\delta \in \mathbb{R}^3$ as an independent vector.
Then, we find the radial basis function weights $w_k$ from Equation~\textcolor{cvprblue}{4} to fit the optimized local displacements $\delta_\text{opt}$.
Here, the basis functions are fit by minimizing the following equation~\cite{tps_1,tps_2},
\begin{align}
    \label{eq:supp_l2}
    & \min_{w} \sum_{\delta_\text{opt}} \|d_w(\mathbf{x}; P_\text{RoI}) - \delta_\text{opt}\|^2 \\ 
    \label{eq:supp_reg}
    & + \lambda {\iint \sum_{i,j}}(\frac{\partial^2 d_w}{\partial x_i \partial x_j})^2 dx_i dx_j,
\end{align}
where $d_w(\mathbf{x}; P_\text{RoI}) = \sum_k w_k\phi(\|\mathbf{x} - \mathbf{p_k}\|)$ is the local displacement map.
To ensure smooth maps are resilient to noise from outliers, we incorporate a regularization term (Equation~\ref{eq:supp_reg}) with a weighting parameter $\lambda$ set to 0.5.

\section{Additional Experimental Results}
\subsection{Comparison with Additional Baselines}
\begin{table}[t]
    \centering
    \resizebox{\linewidth}{!}{
    \begin{tabularx}{0.7\textwidth}{l|CC|CC}
    \toprule
    Metric & \multicolumn{2}{c|}{Bijectivity PCP} & \multicolumn{2}{c}{Chamfer Acc.} \\ \midrule
    Threshold & 0.25 & 0.50 & 0.15 & 0.20 \\ \midrule
    Multi-view Semantic Corresp. (DeiT III~\cite{deit_3}) & 0.05 & 0.09 & 0.24 & 0.45 \\
    Multi-view Semantic Corresp. (MAE~\cite{mae}) & 0.04 & 0.08 & 0.23 & 0.45 \\
    Visual Feature Field (DeiT III~\cite{deit_3}) & 0.31 & 0.34 & 0.58 & 0.67 \\
    Visual Feature Field (MAE~\cite{mae}) & 0.50 & 0.56 & 0.70 & \textbf{0.80} \\
    Ours & \textbf{0.70} & \textbf{0.73} & \textbf{0.71} & 0.76 \\
    \bottomrule
    \end{tabularx}
    }
    \caption{3D scene analogy comparison in manually collected scene pairs in 3D-FRONT~\cite{td_front}. We test baselines with additional vision foundation model features (DeiT III~\cite{deit_3}, MAE~\cite{mae}).}
    \label{tab:td_front_add_baseline}
\end{table}

We compare our method with additional baselines using vision foundation models.
Recall in Section~\textcolor{cvprblue}{4.1} we design the multi-view semantic correspondence~\cite{el2024probing,dinov2} and visual feature field~\cite{sparsedff} baselines that exploit DINOv2~\cite{dinov2} features for finding scene analogies.
Here, we consider additional baselines using different vision foundation models, namely MAE~\cite{mae} and DeiT III~\cite{deit_3}.
Table~\ref{tab:td_front_add_baseline} shows the accuracies of predicted scene analogies in the manually collected scene pairs in 3D-FRONT~\cite{td_front}.
Our method constantly outperforms the newly added baselines, suggesting the effectiveness of the learned descriptor field features for scene context reasoning.

\subsection{Ablation on Semantic and Distance Embeddings}
\begin{table}[t]
    \centering
    \resizebox{0.85\linewidth}{!}{
    \begin{tabularx}{.64\textwidth}{l|CC|CC}
    \toprule
    Metric & \multicolumn{2}{c|}{Bijectivity PCP} & \multicolumn{2}{c}{Chamfer Acc.} \\ \midrule
    Threshold & 0.25 & 0.50 & 0.15 & 0.20 \\ \midrule
    Ours w/o Semantic Emb. & 0.50 & 0.55 & 0.68 & \textbf{0.81} \\
    Ours w/o Distance Emb. & 0.67 & 0.70 & 0.69 & 0.76 \\
    Ours & \textbf{0.70} & \textbf{0.73} & \textbf{0.71} & 0.76 \\
    \bottomrule
    \end{tabularx}
    }
    \vspace{-0.5em}
    \caption{Ablation study of the semantic and distance embeddings on manually collected scene pairs from 3D-FRONT~\cite{td_front}. We report the metric values at varying thresholds.}
    \label{tab:ablation_supp}
\end{table}

We conduct an additional ablation on using semantic and distance embeddings for scene analogy estimation.
Recall in Section~\textcolor{cvprblue}{3.2} our context descriptor fields aggregate semantics and distance information of keypoints near the query point and produces semantic and distance embeddings.
Table~\ref{tab:ablation_supp} shows the scene map accuracy measured from manually collected scenes in 3D-FRONT~\cite{td_front}, where optimal performance is achieved when both are used as input.
Omitting semantic embeddings results in a large performance drop, highlighting the importance of encoding nearby semantic information for effective scene analogy estimation.

\subsection{Quantitative Evaluation of Sim2Real Map Estimation}
\begin{table}[t]
    \centering
    \resizebox{0.6\linewidth}{!}{
    \begin{tabularx}{0.4\textwidth}{l|CC}
    \toprule
     & \begin{tabular}[c]{@{}c@{}}3D Point\\ Feature Field\end{tabular} & Ours \\ \midrule
    Chamfer Acc. & 0.55 & \textbf{0.66} \\ \bottomrule
    \end{tabularx}
    }
    \caption{Quantitative evaluation of 3D scene analogy estimation in manually collected Sim2Real scene pairs from 3D-FRONT~\cite{td_front} and ARKitScenes~\cite{arkit} We report the Chamfer Accuracy at threshold 0.15.}
    \label{tab:sim2real_quant}
\end{table}

We conduct a quantitative evaluation of Sim2Real map estimation using 102 manually collected Sim2Real scene pairs from 3D-FRONT~\cite{td_front} and ARKitScenes~\cite{arkit}.
We compare our method against the 3D Point Feature Field baseline, which is the strongest performing baseline in Tables~\textcolor{cvprblue}{1, 2}.
As shown in Table~\ref{tab:sim2real_quant}, our method outperforms the 3D Point Feature Field baseline in Chamfer accuracy at threshold 0.15.

\subsection{Runtime Characteristics}
We report the runtime for finding 3D scene analogies using neural contextual scene maps.
As our method operates using sparse keypoints, affine and local displacement map estimation can quickly run on average 0.67s and 0.57s, respectively.

\subsection{Robustness Evaluation on Object Groups with Different Cardinalities}
\begin{figure}[t]
  \centering
    \includegraphics[width=\linewidth]{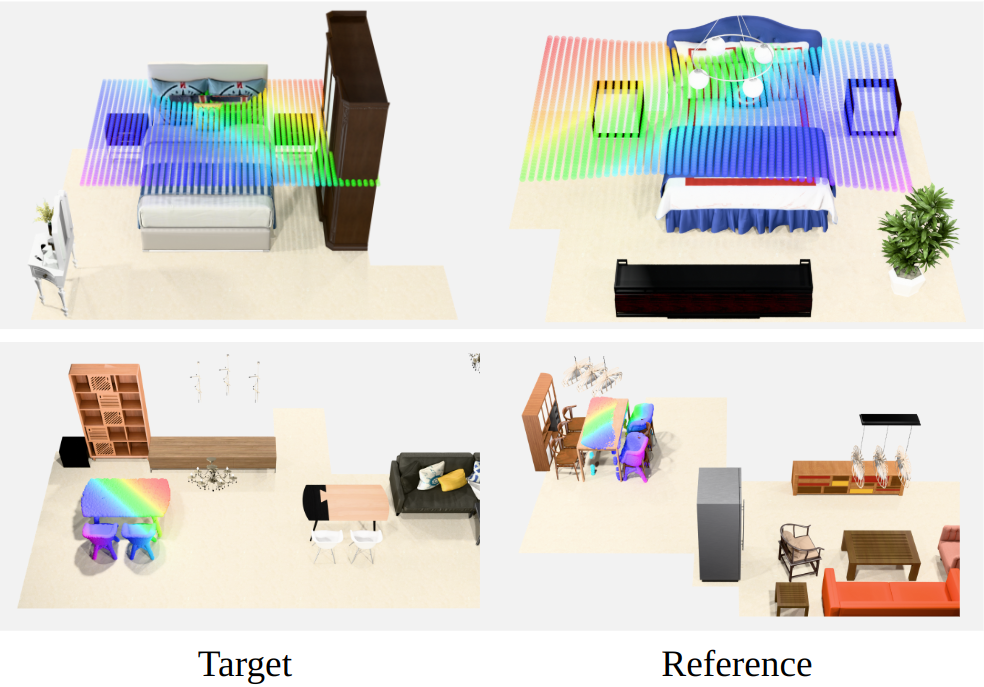}
   \caption{Qualitative results of scene analogies found from our method on object groups with varying cardinalities. We show results both for near-surface and open-space points.}
   \label{fig:object_outlier}
\end{figure}
\begin{table}[t]
    \centering
    \resizebox{0.7\linewidth}{!}{
    \begin{tabularx}{0.4\textwidth}{l|CCC}
    \toprule
     Cardinality & Identical & Different & Original \\ \midrule
    Chamfer Acc. & \textbf{0.76} & 0.66 & 0.71 \\ \bottomrule
    \end{tabularx}
    }
    \caption{Quantitative evaluation of 3D scene analogy estimation in object groups with identical or different cardinalities. We report the Chamfer Accuracy at threshold 0.15 for manually collected scene pairs in 3D-FRONT~\cite{td_front}. Compared to the original metric reported in Table~\textcolor{cvprblue}{1}, our method shows consistent performance amidst object group cardinality variations.}
    \label{tab:object_outlier}
\end{table}
Due to the outlier rejection explained in Section~\ref{sec:affine_supp} and Section~\textcolor{cvprblue}{3.3}, our method can robustly estimate scene analogies in scenarios where object group cardinalities differ.
As shown in Figure~\ref{fig:object_outlier}, our method can estimate scene analogies for cases when (i) the RoI includes objects not present in the reference scene (Figure~\ref{fig:object_outlier} top) and (ii) the reference scene includes objects not present in the target scene (Figure~\ref{fig:object_outlier} bottom).
We further report the accuracy of the estimated maps for scene pairs with object groups having identical / different cardinalities in Table~\ref{tab:object_outlier}, where our method performs constantly in both cases.

\subsection{Performance Analysis with Respect to the Number of RoI Points}
\begin{table}[t]
    \centering
    \resizebox{\linewidth}{!}{
    \begin{tabularx}{.6\textwidth}{l|CCCC}
    \toprule
    \multirow{2}{*}{\begin{tabular}[c]{@{}l@{}}Metric\\ (Points Sampled per Object)\end{tabular}} & \multirow{2}{*}{\begin{tabular}[c]{@{}c@{}}PCK\\ (50)\end{tabular}} & \multirow{2}{*}{\begin{tabular}[c]{@{}c@{}}PCK\\ (100)\end{tabular}} & \multirow{2}{*}{\begin{tabular}[c]{@{}c@{}}PCK\\ (200)\end{tabular}} & \multirow{2}{*}{\begin{tabular}[c]{@{}c@{}}PCK\\ (400)\end{tabular}} \\
    &  &  &  &  \\ \midrule
    Scene Graph Matching & 0.23 & 0.25 & 0.26 & 0.26 \\
    Multi-view Semantic Corresp. & 0.09 & 0.09 & 0.10 & 0.10 \\
    Visual Feature Field & 0.44 & 0.48 & 0.49 & 0.50 \\
    3D Point Feature Field & 0.50 & 0.54 & 0.55 & 0.56 \\
    Ours & \textbf{0.68} & \textbf{0.73} & \textbf{0.75} & \textbf{0.76} \\
    \bottomrule
    \end{tabularx}
    }
    \caption{Quantitative comparison of scene analogies in the procedurally generated scene pairs from 3D-FRONT~\cite{td_front}. We measure percentage of correct points (PCP) at threshold 0.25 using varying number of points samples from the region of interest $P_\text{RoI}$. Compared to the PCK metric measured with 400 points sampled per object (which is mainly used for the experiments), our method performs stably amidst varying number of point samples.}
    \label{tab:dense_pcp}
\end{table}
As specified in Section~\ref{sec:supp_metric_details}, we sample 400 points per each object in the RoI for estimating and evaluating scene analogies.
In this section we evaluate map estimation performance with respect to the number of RoI points.
As shown in Table~\ref{tab:dense_pcp}, our method constantly outperforms the baselines under RoI point variations.
By holistically aligning descriptor fields using smooth maps, our method can attain robustness against individual point locations or point sampline rates and exhibit consistent performance. 

\subsection{Long Trajectory Transfer Comparison}
\begin{figure*}[t]
  \centering
    \includegraphics[width=\linewidth]{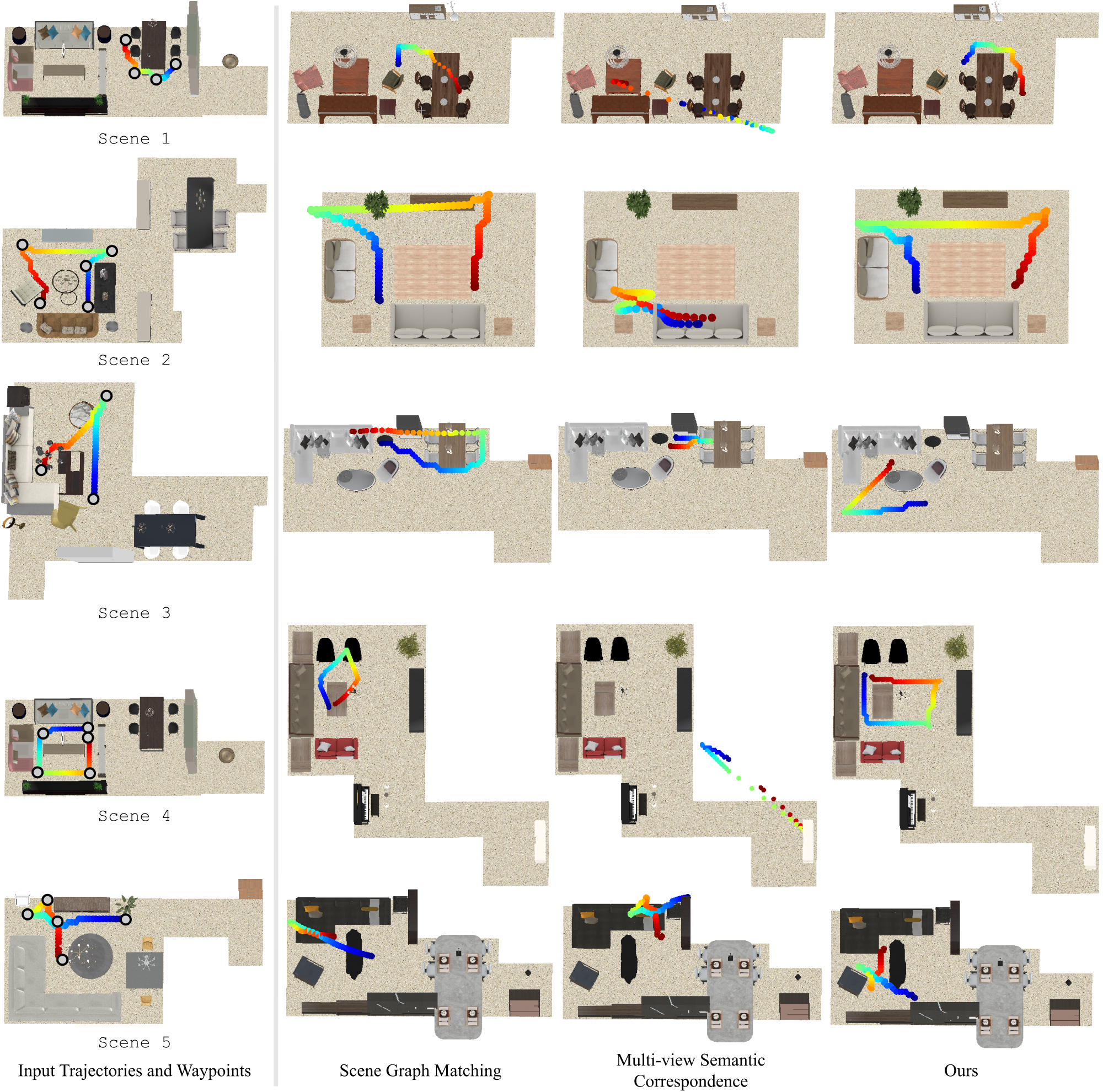}
   \caption{Long trajectory transfer comparison against the scene graph matching and multi-view semantic correspondence baselines in 3D-FRONT~\cite{td_front}.}
   \label{fig:traj_comp}
\end{figure*}

\begin{figure*}[t]
  \centering
    \includegraphics[width=\linewidth]{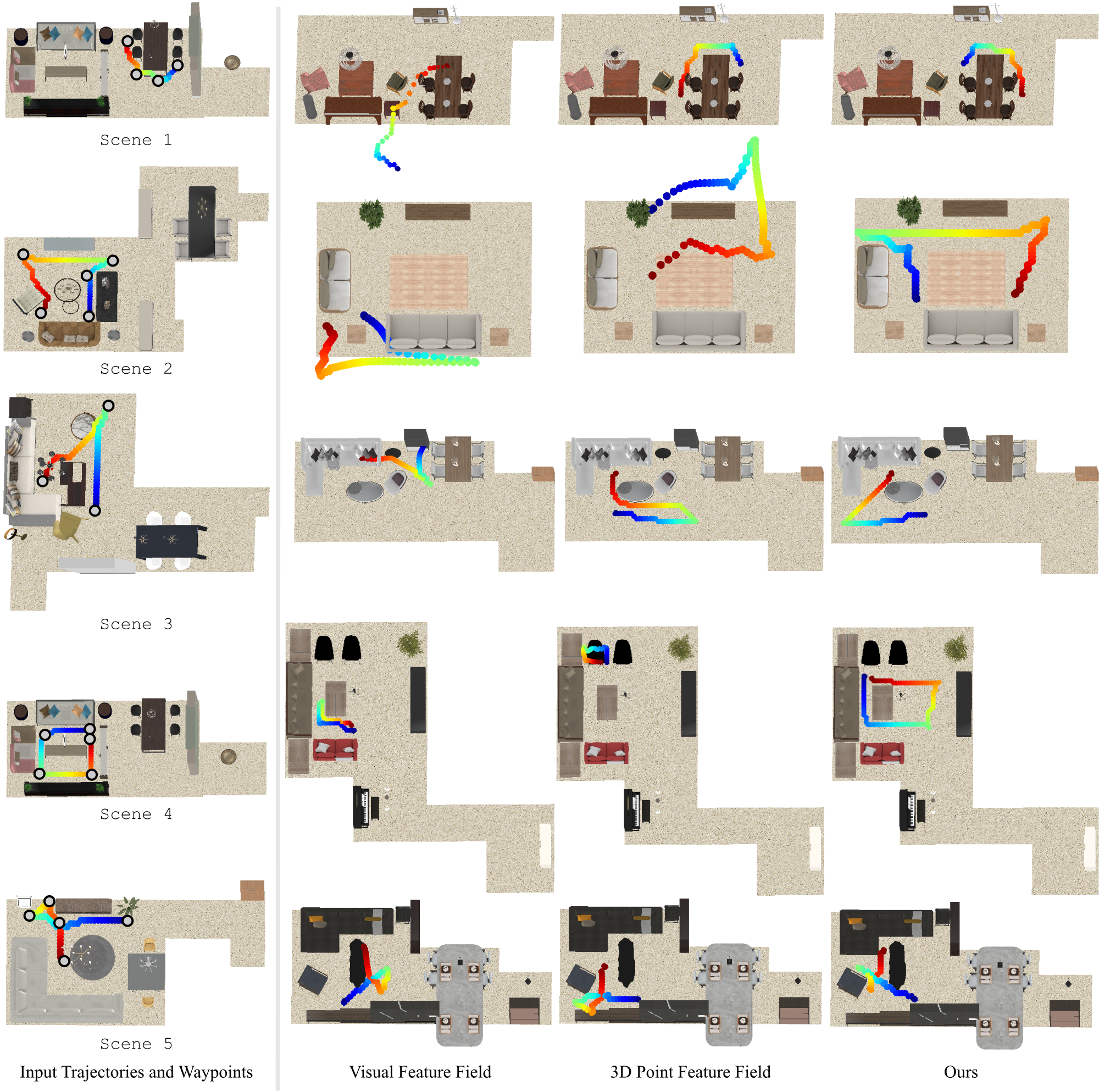}
   \caption{Long trajectory transfer comparison against the field alignment-based baselines (visual feature field, 3D point feature field) in 3D-FRONT~\cite{td_front}.}
   \label{fig:traj_comp_field}
\end{figure*}

In Figure~\ref{fig:traj_comp} and Figure~\ref{fig:traj_comp_field} we compare our method against the baselines in long trajectory transfer explained in Section~\textcolor{cvprblue}{4.2}.
Recall to prevent collisions from directly applying scene maps on long trajectories, we proposed selectively mapping waypoints and interpolating the transferred waypoints via classical path planning (in our case the A* algorithm~\cite{astar}).
Note the scene maps are obtained by setting the objects near the waypoints as the region of interest.
For baselines that only output object surface point matches (scene graph matching, multi-view semantic correspondence), we interpolate object surface matches from the baselines to open space using thin plate spline interpolation~\cite{tps_1,tps_2} and find waypoint transfers.
For cases where the A* algorithm fails to find a path due to inaccurate waypoint transfer, we directly use the interpolated map to transfer short trajectory fragments formed from the failed set of waypoints.
A similar approach is taken for field alignment-based methods (visual feature field, 3D point feature field), while we skip the interpolation process as the output is already a continuous map.

As shown in Figures~\ref{fig:traj_comp} and ~\ref{fig:traj_comp_field}, our method can accurately place waypoints to the coherent location in the reference scene, resulting in long trajectory transfers respecting scene context.
For example, our method can preserve the loop structure in \texttt{Scene 4} or the ribbon-like structure in \texttt{Scene 5} while placing all the waypoints at contextually similar locations.
On the other hand, the baselines often fail to perform appropriate waypoint transfer, resulting in penetrations or misplacements of the transferred trajectory.
Based on the descriptor field that distinguishes contextual information from geometry and semantics, our method can effectively handle waypoint transfers in various scenes.

\begin{figure}[t]
  \centering
    \includegraphics[width=\linewidth]{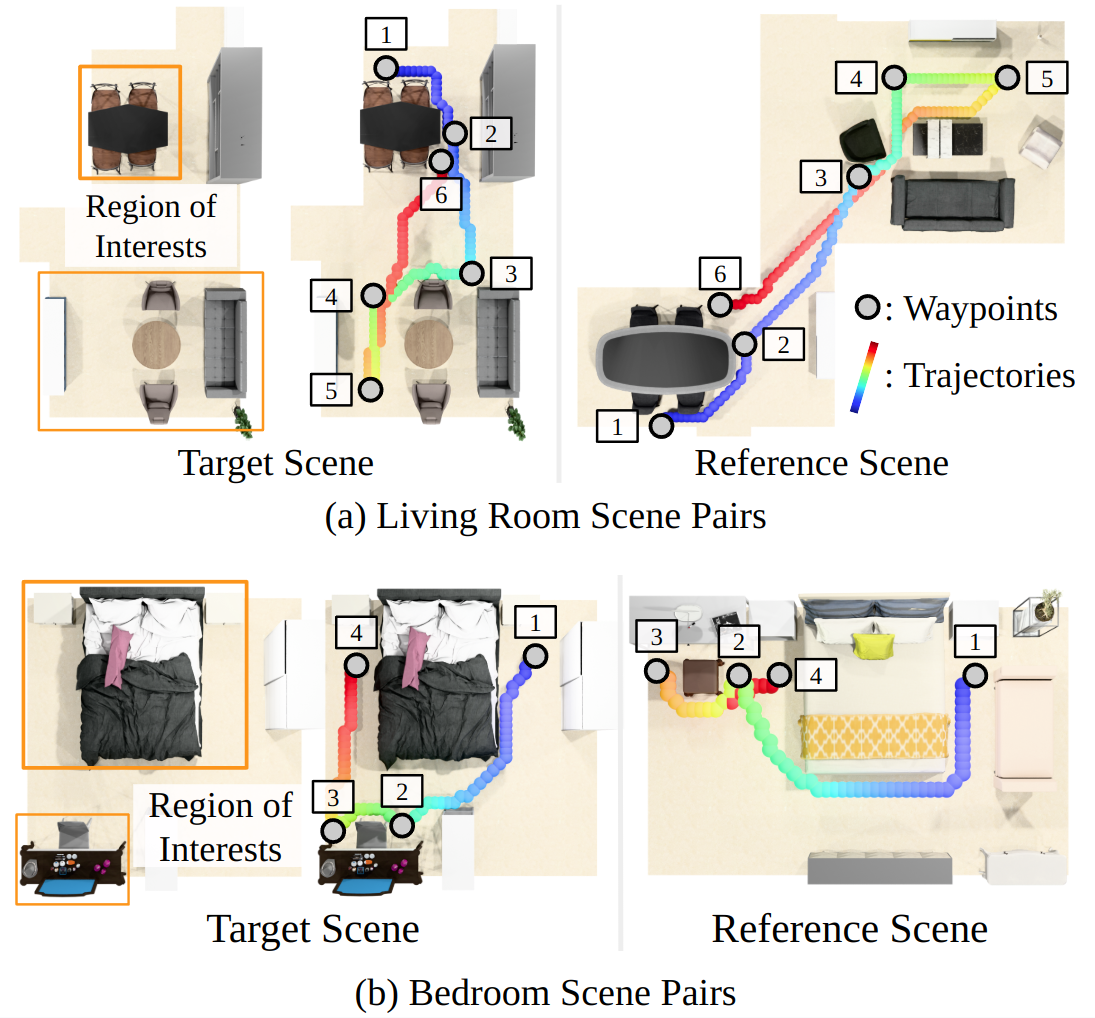}
   \caption{Visualization of long trajectory transfer on 3D-FRONT~\cite{td_front} scene pairs using scene analogies from multiple regions of interest. We use the estimated maps to transfer waypoints, and apply traditional path planning~\cite{astar} to obtain long trajectories spanning the entire 3D scene. We denote the waypoints as gray dots, and the estimated trajectories as color-coded spheres.}
   \label{fig:waypoint_transfer_long}
\end{figure}

\subsection{Trajectory Transfer Using Multiple Regions of Interest}
\label{sec:multi_roi_transfer}
In this section we demonstrate the possibility of using our method for transferring trajectories by using multiple regions of interest.
For long trajectories where a single scene analogy may be difficult to find, our method can instead transfer a sparse set of waypoints and use classical path planning~\cite{astar} for interpolation.
Given a target scene segmented into multiple RoIs as shown in Figure~\ref{fig:waypoint_transfer}, we set waypoints as sampled points in the input trajectory within each RoI.
Note such coarse segmentations can be performed using scene graph clustering~\cite{scene_graph_clustering_1,open3dsg} or vision language models~\cite{molmo2024,flamingo,topviewrs}.

We then find scene analogies for \textit{multiple} RoIs and holistically align them.
To account for symmetry ambiguities (e.g., table + 4 chair RoI in Figure~\ref{fig:waypoint_transfer}), for each RoI we have our method to output the top-5 maps with the smallest cost (Equation~\textcolor{cvprblue}{6}), which results in \textit{combinations} of scene maps.
Note we still apply the validity threshold $\rho_\text{valid}$ explained in Section~\textcolor{cvprblue}{3.3} to filter invalid mappings, which results in a relatively small number of mappings per RoI.
Given a scene with $N_\text{RoI}$ number of RoIs, this procedure results in at most  $5^{N_\text{RoI}}$ possible \textit{combinations} of mappings.

Next, we choose the optimal combination via a simple criterion based on isometry preservation~\cite{functional_maps,functional_maps_2}.
Here we take inspiration from prior works in 3D surface mapping~\cite{functional_maps,functional_maps_2} that often impose isometry constraints such that the local geometric structure is preserved under non-rigid deformations.
To elaborate, let $P_\text{rand} \in \mathbb{R}^{N_\text{rand} \times 3}$ be a set of randomly sampled points from the multiple regions of interest, and the distance matrix $D_\text{rand} \in \mathbb{R}^{N_\text{rand} \times N_\text{rand}}$ whose $(i, j)^\text{th}$ entry contains the euclidean distance between the $i^\text{th}$ and $j^\text{th}$ points.
Similarly, for an arbitrary map combination, let $P_\text{transform}$ be the transformation result of $P_\text{rand}$ under the map combination and $D_\text{transform}$ the distance matrix.
The isometry cost is then defined as the Frobenius norm between the distance matrices, namely $\|D_\text{rand} - D_\text{transform}\|_F$.
We aim to find the combination with a small isometry cost, where we employ a simple greedy approach.
Given a randomly initialized combination, we sequentially update the map associated with each RoI to the one that produces a smaller isometry cost among the top-5 (or lower due to filtering) estimated maps.
This process is repeated for a fixed number of iterations, and we use the final map combination to produce long trajectory transfers.
While the search process is quite simple, we find this method to work well for scenes with a moderate number of RoI segments ($N_\text{RoI} < 5)$.

Finally, we transfer each waypoint using the mapping found for the associated RoI, and interpolate between the transferred waypoints using classical path planning~\cite{astar}.
As shown in Figure~\ref{fig:waypoint_transfer}, the proposed method can align multiple scene analogies and produce a coherent long trajectory transfer spanning over the entire 3D scene.
Nevertheless, devising a more scalable and principled approach to align multiple scene maps originating from different RoIs is left as future work.

\section{Experimental Setup Details}

\begin{figure}[t]
  \centering
    \includegraphics[width=\linewidth]{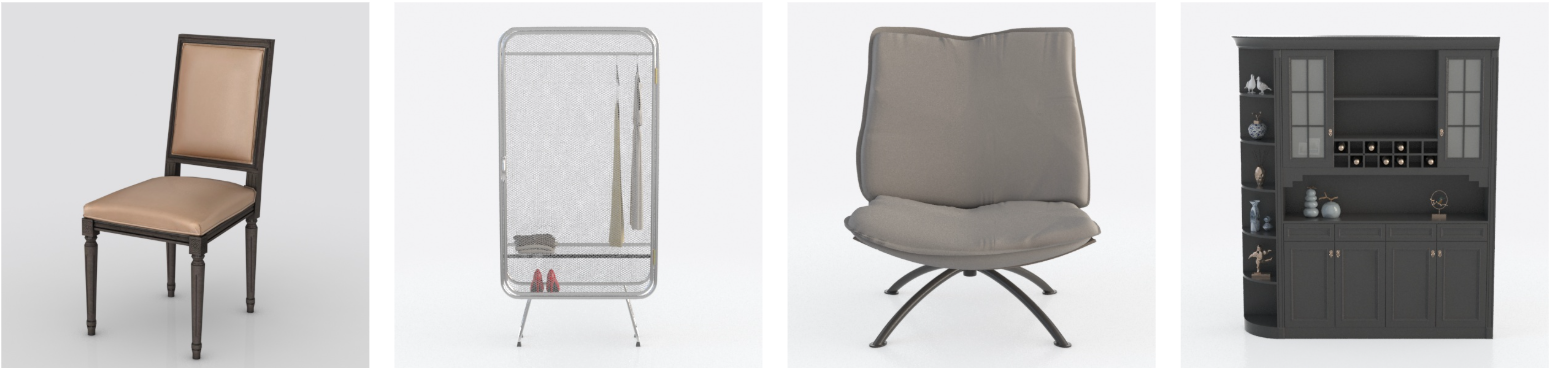}
   \caption{Frontal view renderings of objects in 3D-FRONT~\cite{td_front}, used for CLIP~\cite{clip} and sentence embedding~\cite{sentence_bert} feature extraction.}
   \label{fig:supp_td_front_render}
\end{figure}

\subsection{Baselines}
In this section, we elaborate on the implementation details of the baselines compared against our method.
As the 3D scene analogy task is new, we tailor existing 3D scene understanding pipelines to our task and introduce four baseline approaches capable of outputting dense scene maps.

\paragraph{Scene Graph Matching}
The scene graph matching baseline builds a 3D scene graph~\cite{td_scene_graph_armeni} representing each object as nodes and finds affine transformations to align the graphs.
First, we build scene graphs in a rule-based manner following Jia et al~\cite{jia2024sceneverse}, where we use object bounding box intersections to determine scene graph edge types.
Then, given a pair of 3D scene graphs $\mathcal{G}_\text{tgt}, \mathcal{G}_\text{src}$ for the target and reference scene, we list all subgraphs in $\mathcal{G}_\text{ref}$ and compare them against the subgraph containing the region of interest in $\mathcal{G}_\text{tgt}$.
Here, we measure similarities between subgraphs using the Jaccard coefficient introduced by Wald et al~\cite{td_ssg}.
After retrieving the closest subgraph in $\mathcal{G}_\text{ref}$ to the region of interest, we apply Hungarian matching~\cite{sg_pgm,hungarian} between the subgraph nodes by using object semantic labels and adjacent edge labels as node features.
Finally, similar to Sarkar et al.~\cite{sgaligner}, we find an affine transformation from the node matches and deduce the final point-level alignment by performing iterative closest points (ICP)~\cite{icp} separately for object point clouds associated with each node match.

\paragraph{Multi-view Semantic Correspondence}
The multi-view semantic correspondence baseline renders scenes at multiple views and operates based on 2D matches from vision foundation models~\cite{el2024probing}.
To elaborate, we sample $N_\text{render}=5$ views from virtual spheres encompassing $S_\text{tgt}$ and $S_\text{ref}$~\cite{wang2021neus,yariv2021volume}, and extract DINOv2~\cite{dinov2} features for each view.
Then, we exhaustively match $N_\text{render}^2$ image pairs using the extracted features~\cite{el2024probing}, and lift each 2D match to 3D via back-projection.
Using the 3D matches, we obtain object-level matches by having each 3D match vote for an object pair.
In this phase, for each object in the region of interest, the object in the reference scene with the largest amount of votes is selected.
As the last step, we estimate affine transforms using the matched object centroids followed by iterative closest points (ICP)~\cite{icp} to get the point-level alignments.
Note that while it is possible to directly use the 3D matches and interpolate them to get point-level matches, we find the DINOv2~\cite{dinov2} descriptors to be quite noisy for obtaining fine-grained matches between object groups.
Therefore, our baseline implementation mainly uses the features for object-level matching, which we empirically find to be more effective.

\paragraph{Visual Feature Field}
Instead of lifting 2D matches, the visual feature field baseline directly finds smooth scene maps by aligning vision foundation model features in 3D.
The baseline first renders $N_\text{render}=5$ views from virtual spheres encompassing the input scenes, and extracts DINOv2~\cite{dinov2} features.
Next, the baseline computes multi-view aggregated features at each 3D keypoint in $S_\text{tgt}$ and $S_\text{ref}$.
Here, the method projects each 3D keypoint to the rendered views and extracts keypoint features via bilinear interpolation, and averages the $N_\text{render}$ features.
For an arbitrary query point, we compute features by using distance-weighted interpolation as in Wang et al.~\cite{sparsedff,dff}.
In this step, we aggregate features by considering keypoints within radius $r$ from the query point, where the radius values are set identical to our method.
Finally, the baseline applies the coarse-to-fine map estimation from Section~\textcolor{cvprblue}{3.3} to obtain scene analogies.

\paragraph{3D Point Feature Field}
Similar to the visual feature field baseline, the 3D point feature field interpolates keypoint features to obtain features at arbitrary locations, but uses 3D keypoint descriptors~\cite{vector_neuron} instead of vision foundation models.
For feature extraction, we use PointNet~\cite{pointnet,pointnet_pp} containing Vector Neuron layers~\cite{vector_neuron} that is pre-trained on the ModelNet40~\cite{modelnet} dataset.
Here we use the rotation-invariant embeddings obtained from the last layer of the Vector Neuron~\cite{vector_neuron} encoder prior to max pooling.
Given the 3D keypoint features, we obtain features at arbitrary locations via distance-weighted interpolation~\cite{sparsedff}, and align the keypoint-based fields using our coarse-to-fine estimation scheme.

\subsection{Foundation Model Features for Ablation Study}
In Section~\textcolor{cvprblue}{4.1.2}, we demonstrate scene analogy estimation using vision and language foundation model features, namely CLIP~\cite{clip} and sentence embeddings~\cite{bert,sentence_bert}.
Here we elaborate on the details of the experiment.
For CLIP feature extraction, we first render frontal views of 3D-FRONT~\cite{td_front} objects as shown in Figure~\ref{fig:supp_td_front_render} and extract CLIP feature embeddings.
The embeddings are then used in place of the semantic label embedding introduced in Section~\textcolor{cvprblue}{3.2}.
For sentence embedding extraction, we first apply an off-the-shelf image captioning method on each of the object renderings in 3D-FRONT~\cite{td_front}.
Then, we extract sentence embeddings for each of the image captions, and supply them as input to the descriptor fields in place of the semantic embeddings.

\begin{figure}[t]
  \centering
    \includegraphics[width=\linewidth]{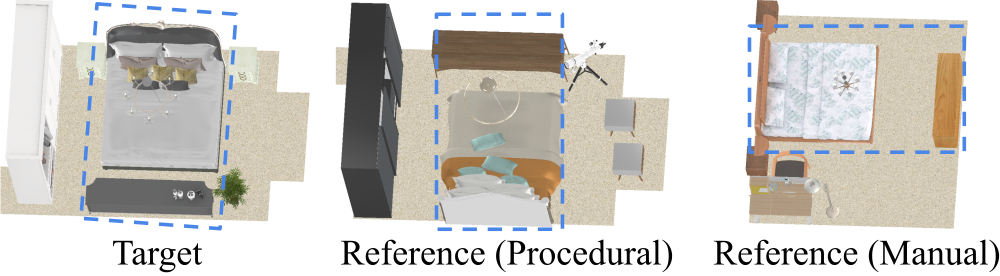}
   \caption{Qualitative sample of scene pairs for evaluation. The blue box denotes the common object groups present both in the target and reference scenes.}
   \label{fig:eval_scene_pairs}
\end{figure}
\begin{figure*}[t]
  \centering
    \includegraphics[width=\linewidth]{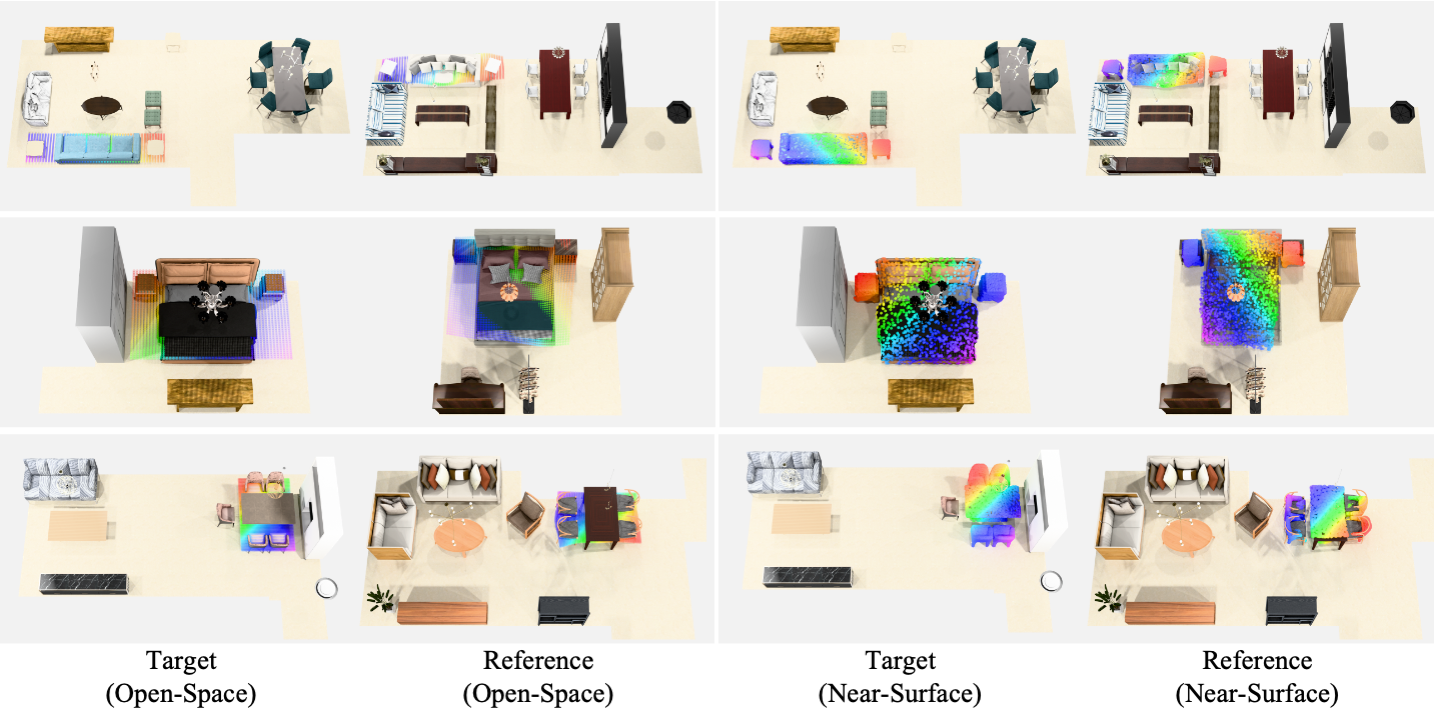}
   \caption{Additional visualization of 3D scene analogies estimated in 3D-FRONT~\cite{td_front}. We show results both for near-surface and open-space points.}
   \label{fig:supp_td_front}
\end{figure*}
\begin{figure*}[t]
  \centering
    \includegraphics[width=\linewidth]{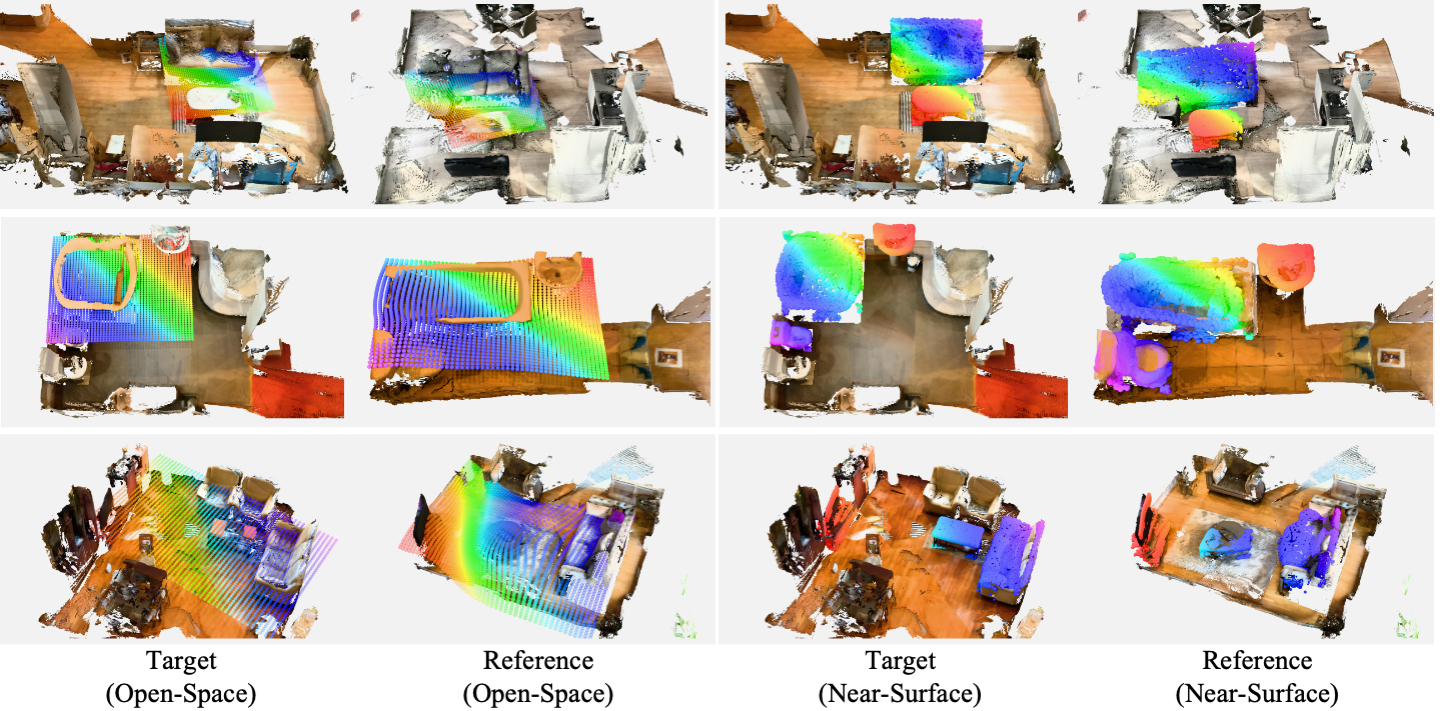}
   \caption{Additional visualization of 3D scene analogies estimated in ARKitScenes~\cite{arkit}. We show results both for near-surface and open-space points.}
   \label{fig:supp_arkit}
\end{figure*}

\subsection{Scene Pair Preparation for Evaluation}
We elaborate on the scene pair preparation process for evaluating scene analogies in Section~\textcolor{cvprblue}{4.1}.
As shown in Figure~\ref{fig:eval_scene_pairs}, we prepare two types of data, namely procedurally generated and manually collected scene pairs.

\paragraph{Procedurally Generated Scene Pairs}
Recall these scene pairs contain pseudo ground-truth annotations for evaluating point-level accuracy of scene analogy estimations.
For each scene, we first randomly select an object and its k-nearest neighbors (where k is randomly sampled from $\{2,\dots,4\}$).
The points sampled from the selected objects are used as the region of interest $P_\text{RoI}$.
Then, for objects not selected, we either randomly remove them by a probability of 0.5 or apply pose perturbation.
Here translation noise is sampled from the uniform distribution $\mathcal{U}(-0.05, 0.05)$ and rotation noise is obtained from the set of z-axis rotations with rotation angles sampled from $\mathcal{U}(-10^\circ, 10^\circ)$.
Next, we randomly add $N_\text{add}$ objects to open spaces in the scene (where $N_\text{add}$ is sampled from $\mathcal{U}(2, 5)$).
In this step, we retrieve the scene in the evaluation dataset with the closest object semantic label histogram, and select objects from that scene for addition.
The objects are added by computing an occupancy grid map of the current scene and randomly choosing from collision-free locations~\cite{occ_grid_map}.
Finally, we replace each object that has not been added or removed during the previous steps with a randomly selected object of the same semantic class, similar to training data generation explained in Section~\ref{sec:supp_context_desc}.
The resulting procedurally generated scenes contain realistic object placements while preserving meaningful object group structures for evaluation.

To compute pseudo ground-truth scene analogies for $P_\text{RoI}$, we uniformly sample points from the matching object group in the procedurally generated scene.
Then, we apply Hungarian matching~\cite{hungarian} between the two point sets, which yields an injective matching for each point in $P_\text{RoI}$ to the sampled points in the generated scene.
We use this matching result as the pseudo ground-truth for evaluation.
Using the entire process, we generate 997 scene pairs for 3D-FRONT~\cite{td_front} and 549 scene pairs for ARKitScenes~\cite{arkit}.

\paragraph{Manually Collected Scene Pairs}
In addition to the procedurally generated pairs, we manually collect scene pairs for evaluation.
As obtaining point-level manual annotations is costly and possibly inaccurate, we only make group-level annotations for scene pairs.
Specifically, for each scene pair sharing common object groups, we annotate the instance IDs of objects within the groups.
We further annotate scene pairs not containing any common object groups, which we use for checking false positive scene analogies.
We collect 120 scene pairs containing 20 pairs having no object group matches for both 3D-FRONT~\cite{td_front} and ARKitScenes~\cite{arkit}.

\begin{figure}[t]
  \centering
    \includegraphics[width=\linewidth]{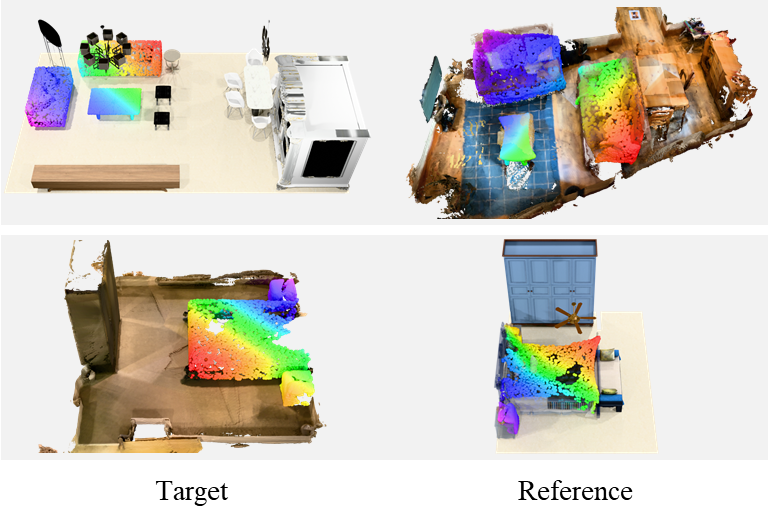}
   \caption{Additional visualization of Sim2Real and Real2Sim scene analogies estimated between 3D-FRONT (Sim) and ARKitScenes (Real).}
   \label{fig:supp_sim2real}
\end{figure}
\begin{figure}[t]
  \centering
    \includegraphics[width=\linewidth]{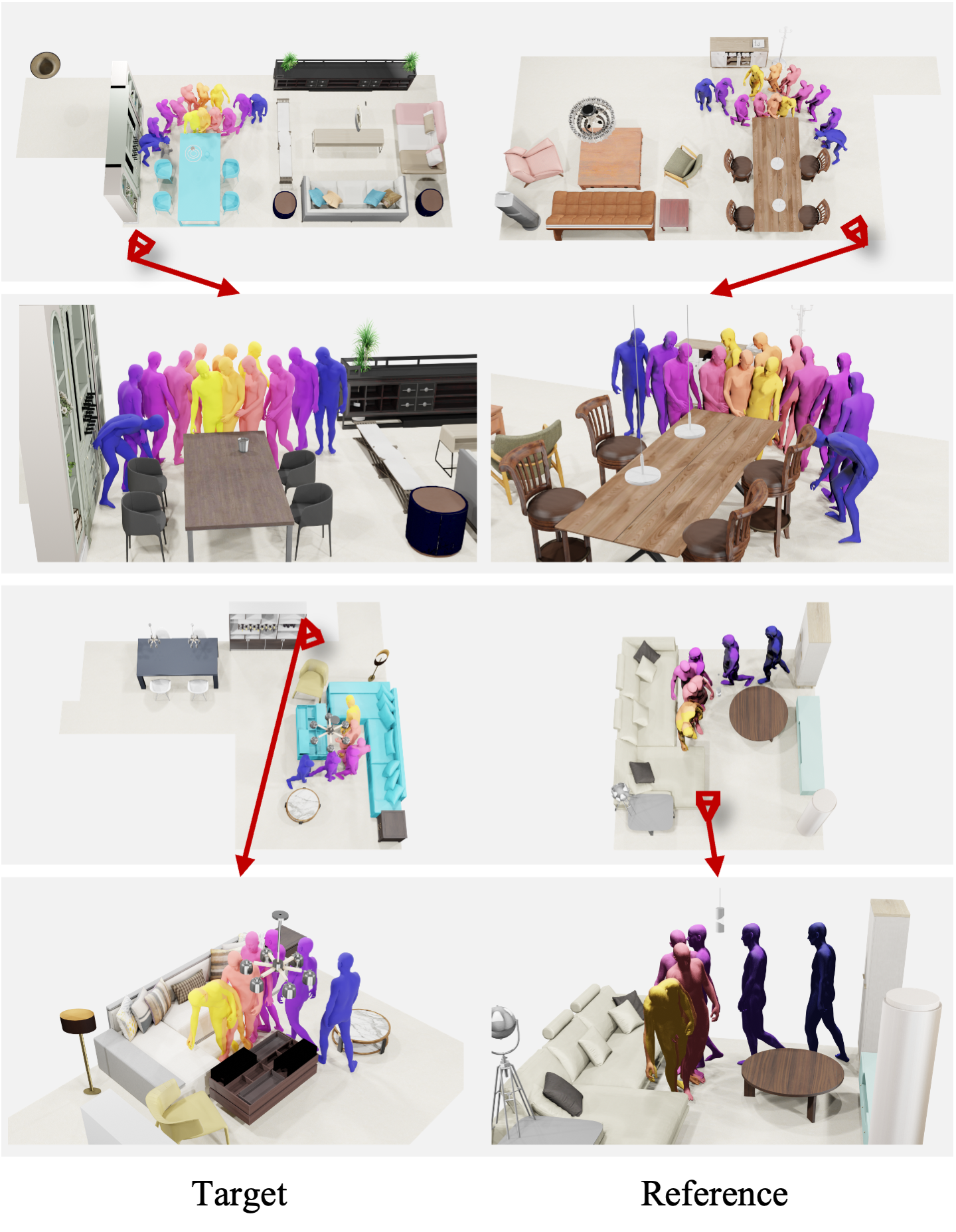}
   \caption{Additional visualization of short trajectory transfer. We shade the region of interest used for estimating scene analogies in blue.}
   \label{fig:supp_trajectory}
\end{figure}
\begin{figure}[t]
  \centering
    \includegraphics[width=\linewidth]{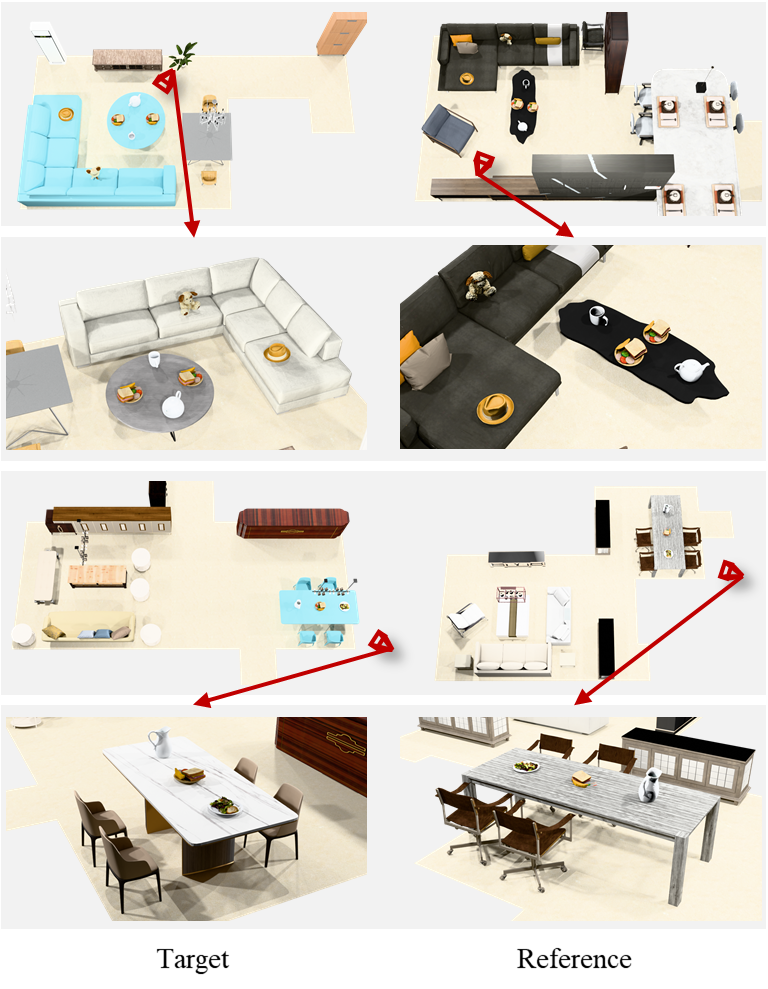}
   \caption{Additional visualization of object placement transfer. We shade the region of interest used for estimating scene analogies in blue.}
   \label{fig:supp_object}
\end{figure}

\subsection{Evaluation Metric Details}
\label{sec:supp_metric_details}
\paragraph{Percentage of Correct Points (PCP) and Bijectivity PCP}
The percentage of correct points (PCP) metric is measured for procedurally generated scene pairs having pseudo ground-truth annotations to evaluate point-level accuracy of scene maps, while the bijectivity PCP is a similar metric to measure whether the estimated maps are invertible.
Both metrics are defined for points on the region of interest: namely, we sample 400 points from each object point cloud in the original scene using farthest point sampling (FPS)~\cite{fps}.

\paragraph{Chamfer Accuracy}
The Chamfer Accuracy metric evaluates the group-level accuracy of scene analogy predictions while penalizing false positive maps.
Thus the metric additionally provides evaluation on the false positive rates of each method, i.e., whether the method falsely outputs mappings when the region of interest is unmatchable to the reference scene.
In this section, we formally define the metric.
We first define the Chamfer distance for a pair of point sets $X, Y \in \mathbb{R}^3$ as follows
\begin{equation}
    \text{CD}(X, Y) = \sum_{\mathbf{x} \in X} \min_{\mathbf{y} \in Y} \|\mathbf{x} - \mathbf{y}\|_2 + \sum_{\mathbf{y} \in Y} \min_{\mathbf{x} \in X} \|\mathbf{y} - \mathbf{x}\|_2.
\end{equation}
Given the object set $\mathcal{O}_\text{RoI} = \{P_i^\text{RoI}\}$ in the region of interest, we perform nearest neighbor matching using object centroid locations to obtain corresponding objects in the reference scene $\mathcal{O}_\text{match} = \{P_i^\text{match}\}$.
Recall that the region of interest is defined as a union of object group points, namely $P_\text{RoI} = \bigcup\limits_{i} P_i^\text{RoI}$.
For scene pairs containing matchable regions, the Chamfer accuracy is then defined as follows,
\begin{align}
    \text{CA}(P_\text{RoI}) = & \mathbbm{1}[\frac{1}{|\mathcal{O}_\text{RoI}|}\sum_i \text{CD}(P_i^\text{RoI}, P_i^\text{match}) \leq \alpha],
\end{align}
where $\alpha$ is a threshold parameter.
For scene pairs labeled as unmatchable, the Chamfer accuracy is set to $1$ if no maps are produced, and $0$ otherwise.

\section{Limitations and Future Work}
\begin{figure}[t]
  \centering
    \includegraphics[width=0.95\linewidth]{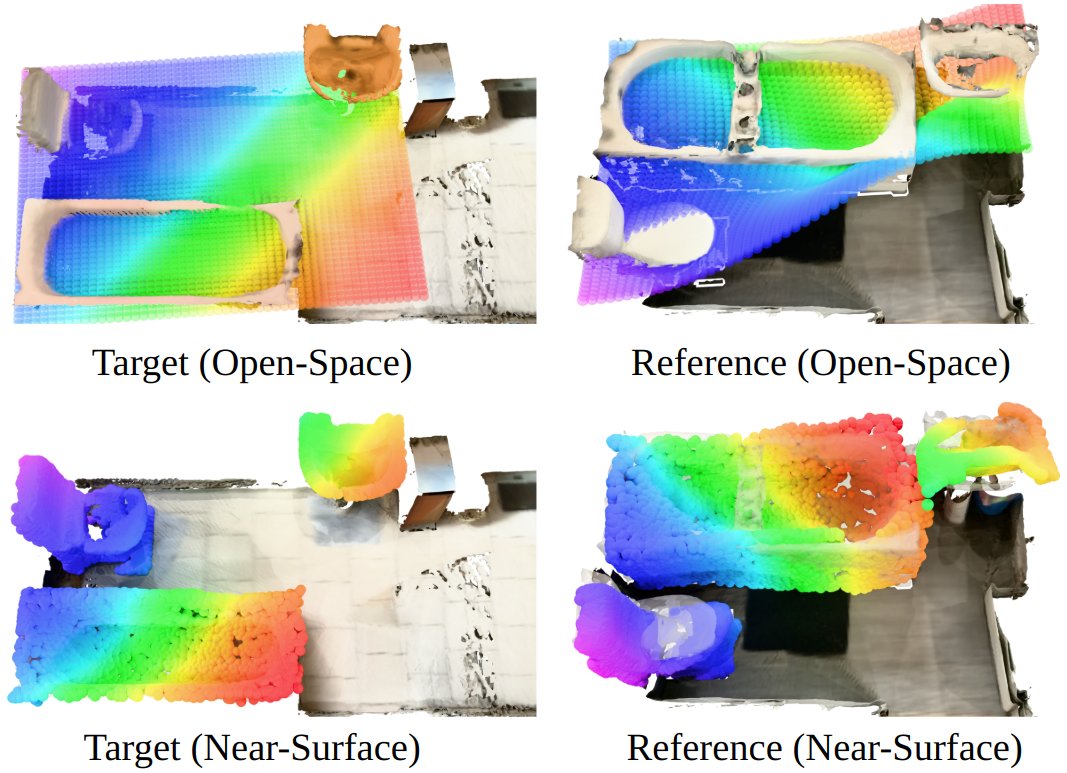}
   \caption{Failure case of our method in scenes where the scene analogy cannot be initially approximated with affine maps, leaded to inaccurate estimations.}
   \label{fig:failure}
\end{figure}

While our new task of finding 3D scene analogies holds practical applications for robotics and AR/VR, and our neural contextual scene maps can effectively find scene analogies, we acknowledge several limitations that invite further investigation in future work. 

\paragraph{Handling Symmetries and Multi-modalities during Evaluation}
We observe reflection symmetries to exist quite often in object groups.
For example, all object groups shown in Figure~\ref{fig:supp_td_front} exhibit such symmetries.
While these groups are symmetric \textit{in isolation}, the ambiguities can be mitigated by leveraging the context from neighboring scene regions.
To illustrate, the symmetric placements of the tables in the first row of Figure~\ref{fig:supp_td_front} can be disambiguated by considering the neighboring objects: one table is next to another sofa, while the other table is next to a group of chairs.
A similar argument can be made for the cabinet-and-bed group in the second row.
Notice that our method correctly recognizes this contextual information and produces maps that respect nearby objects' information.

Nevertheless, there also exist object groups where such disambiguations are not effective: for the third row in Figure~\ref{fig:supp_td_front}, it is unclear from the neighboring scene contexts whether the currently estimated map is the only possible scene analogy.
In this work we take a `lenient' strategy for handling reflection symmetries: we additionally measure the PCP (percentage of correct points) metric for horizontal and vertical reflections, and report the smallest value.
However, we believe symmetries can be better handled by additionally labeling the symmetry type of object groups, for example whether a group is reflection symmetric or can be disambiguated from nearby contexts, similar to how studies in object pose estimation~\cite{scan2cad,rio_wald} evaluate symmetric objects.
Obtaining additional labels and devising better symmetry-aware metrics are left as future work.

In addition to object-wise symmetries, ambiguities can arise in scenes containing higher-level symmetries, namely multiple similar object groups.
While in most cases a single map can unambiguously match object groups, we acknowledge that there are scenarios where multiple scene analogies are detectable.
For example, suppose one wants to find scene analogies between the target scene in Figure~\ref{fig:eval_scene_pairs} and a large room containing multiple bed-and-cabinet combinations.
Although our neural contextual scene maps currently output a \textit{single} mapping, it could be extended in such cases to output the top-K mappings as in Section~\ref{sec:multi_roi_transfer}, which will lead to multiple scene analogy detections.
Nevertheless, devising metrics and annotating scene pairs for multi-modal scene analogies is not straightforward, and thus is open to future work.

\paragraph{Infidelity of Affine Map Estimation}
Although the affine map estimation can effectively handle large, global transformations, we identified cases where the initial affine map estimation failed to find good solutions.
These cases occur when scene analogies between two scenes cannot be approximated with an affine map.
An example is shown in Figure~\ref{fig:failure}, where the relative locations of the toilet and bathtub are swapped, and thus affine maps are insufficient for aligning the scenes.
Our method attempts to find an affine map that best aligns the two scenes, yet errors occur for regions near the sink (observe that the original points are incorrectly mapped to \textit{flipped} regions in the reference scene).
We expect a more flexible set of initializations, for example piece-wise affine transforms~\cite{piece_affine}, can be used in place of the affine mapping procedure to solve such inaccuracies.
Alternatively, finding multiple partial maps (e.g., separately mapping toilet-bathtub and toilet-sink groups for Figure~\ref{fig:failure}) and combinatorially aligning them as in Section~\ref{sec:multi_roi_transfer} could also be a feasible solution.

\paragraph{Scene Pair Generation for Training}
While training descriptor fields does not require densely labeled ground-truth data and descriptor fields can function without semantic labels during inference as demonstrated in Section~\textcolor{cvprblue}{4.1.2}, the training process still requires the generation of positive and negative scene pairs for contrastive learning~\cite{sinclr,simsiam}.
This process demands semantic and instance labels of 3D scenes, along with each object's pose.
Although such information can be reliably extracted from modern 3D segmentation / pose estimation algorithms~\cite{mask3d,yang2023sam3d,xie2020mlcvnet,qi2019deep}, we posit our method to become more scalable if descriptor fields can be learned without exploiting any synthetic scene pairs.
One interesting direction is to distill the knowledge of 3D scene generation methods~\cite{blockfusion,ocal2024sceneteller} trained on large amounts of indoor data for finding 3D scene analogies, similar to how image generation models~\cite{stable_diffusion} have been adapted to semantic correspondence tasks~\cite{tang2023emergent}.
Finding more flexible learning strategies to train descriptor fields is left as future work.

\paragraph{Handling Various Notions of ``Correct" Correspondences}
Inspired from prior works in semantic correspondence~\cite{semantic_corresp_1,semantic_corresp_2,semantic_corresp_3,semantic_corresp_4}, our work considers points having similar nearby object semantics and local geometry to be correct matches, and the descriptor fields are trained to support this notion of ``correctness".
We have demonstrated in Section~\textcolor{cvprblue}{4.2} that this definition is useful for tasks such as trajectory transfer in robotics or object placement transfer in AR/VR.
Nevertheless, we acknowledge that multiple definitions of "correct" correspondences exist depending on the task.
For example, one may want to find scene analogies based on other attributes such as affordance (e.g., mapping `sittable' areas from one scene to another) or appearance (e.g., matching furniture groups with a specific style).
Due to the modular design of our approach of separating descriptor extraction and map estimation based on classical optimization, our method can be flexibly modified to handle such definitions of correctness.
Specifically, one may train new descriptor sets for  different correctness definitions and subsequently apply the map estimation process that does not require training.

\section{Additional Qualitative Results}
We display additional qualitative results for 3D scene analogy estimation in 3D-FRONT~\cite{td_front} (Figure~\ref{fig:supp_td_front}), ARKitScenes~\cite{arkit} (Figure~\ref{fig:supp_arkit}), Sim2Real and Real2Sim (Figure~\ref{fig:supp_sim2real}).
Our method can produce accurate scene maps in all cases, due to the coarse-to-fine estimation framework which enhances robustness against input variations.
We further show additional qualitative results for short trajectory transfer (Figure~\ref{fig:supp_trajectory}) and  object placement transfer (Figure~\ref{fig:supp_object}).
The accurate scene analogy estimations can be effectively exploited for downstream tasks in robotics and AR/VR.

{
    \small
    \bibliographystyle{ieeenat_fullname}
    \bibliography{main}
}


\end{document}